\documentclass[10pt,journal,compsoc,twoside]{IEEEtran}
\ifCLASSOPTIONcompsoc
  \usepackage[nocompress]{cite}
\else
  \usepackage{cite}
\fi
\ifCLASSINFOpdf
   \usepackage[pdftex]{graphicx}
   \graphicspath{{../figures/}}
   \DeclareGraphicsExtensions{.pdf,.jpg}
\else
   \usepackage[dvips]{graphicx}
   \graphicspath{{../figures/}}
   \DeclareGraphicsExtensions{.eps}
\fi
\usepackage{amsmath}
\usepackage{array}
\usepackage{float}

\usepackage{url}

\usepackage{amsfonts}
\usepackage{threeparttable}
\usepackage{xr-hyper}
\usepackage{hyperref}
\usepackage{dcolumn}
\usepackage[dvipsnames]{xcolor}

\newcolumntype{o}{D{.}{.}{1}} 
\newcolumntype{t}{D{.}{.}{3}} 
\newcolumntype{f}{D{.}{.}{4}} 
\newcommand{\dheader}[1]{
    \multicolumn{1}{r}{#1}
}
\makeatletter
\newcolumntype{B}[3]{>{\boldmath\DC@{#1}{#2}{#3}}c<{\DC@end}}
\makeatother
\newcommand{\boldo}[1]{
    \multicolumn{1}{B{.}{.}{1}}{#1}
}
\newcommand{\boldt}[1]{
    \multicolumn{1}{B{.}{.}{3}}{#1}
}
\newcommand{\boldf}[1]{
    \multicolumn{1}{B{.}{.}{4}}{#1}
}

\begin{document}

\bstctlcite{BSTcontrol}
%
\title{Adversarially-regularized mixed effects deep learning (ARMED) models improve interpretability, performance, and generalization on clustered (non-\textit{iid}) data}
%
%
%
%

\author{
        Kevin~P.~Nguyen and
        Albert~A.~Montillo,\\
        {\small for~the~Alzheimer’s~Disease~Neuroimaging~Initiative}
\IEEEcompsocitemizethanks{
    \IEEEcompsocthanksitem The authors were with the Lyda Hill Department
    of Bioinformatics, University of Texas Southwestern Medical Center, Dallas,
    TX, 75390.
    \makeatletter
    E-mail: kevin3.nguyen@utsouthwestern.edu, albert.montillo@utsouthwestern.edu\protect\\
    \makeatother
    \IEEEcompsocthanksitem Data used in preparation of this article were obtained 
    from the Alzheimer’s Disease Neuroimaging Initiative (ADNI) database 
    (adni.loni.usc.edu). As such, the investigators within the ADNI 
    contributed to the design and implementation of ADNI and/or provided data 
    but did not participate in analysis or writing of this report. A complete 
    listing of ADNI investigators can be found at: 
    http://adni.loni.usc.edu/wp-content/uploads/how\_to\_apply/ADNI\_Acknowledgement\_List.pdf
}
}

%
%

\ifCLASSOPTIONpeerreview
    \markboth{Preprint}%
    {ARMED models for clustered data}
\else
    \markboth{Preprint}%
    {Nguyen \& Montillo: ARMED models for clustered data}
\fi
%



\IEEEtitleabstractindextext{%
\begin{abstract}
Natural science datasets frequently violate assumptions of independence. Samples may be clustered (e.g. by study site, subject, or experimental batch), leading to spurious associations, poor model fitting, and confounded analyses. While largely unaddressed in deep learning, this problem has been handled in the statistics community through mixed effects models, which separate cluster-invariant \textit{fixed} effects from cluster-specific \textit{random} effects. We propose a general-purpose framework for Adversarially-Regularized Mixed Effects Deep learning (ARMED) models through non-intrusive additions to existing neural networks: 1) an adversarial classifier constraining the original model to learn only cluster-invariant features, 2) a random effects subnetwork capturing cluster-specific features, and 3) an approach to apply random effects to clusters unseen during training. We apply ARMED to dense, convolutional, and autoencoder neural networks on 4 applications including simulated nonlinear data, dementia prognosis and diagnosis, and live-cell image analysis. Compared to prior techniques, ARMED models better distinguish confounded from true associations in simulations and learn more biologically plausible features in clinical applications. They can also quantify inter-cluster variance and visualize cluster effects in data. Finally, ARMED improves accuracy on data from clusters seen during training (up to 28\% vs. conventional models) and generalization to unseen clusters (up to 9\% vs. conventional models). 
\end{abstract}

}

\maketitle

\IEEEdisplaynontitleabstractindextext

%
\IEEEpeerreviewmaketitle

\IEEEraisesectionheading{\section{Introduction}\label{sec:introduction}}

%
%
%
%
\IEEEPARstart{I}{n} predictive modeling, one often assumes that data is independent and identically distributed (\textit{iid}), such that no samples are correlated or interdependent. However, this assumption is frequently violated in the natural sciences when samples are clustered. For example, many multi-site neurological studies acquire cognitive scores using a different human rater at each site, which are subject to inter-rater differences \cite{Connor.2008, Kozora.2008, Schafer.2011}. As a result, these measurements have inherent intra-site correlation and inter-site variability. Another example is medical imaging, such as magnetic resonance imaging (MRI), where differences in imaging protocol and scanner hardware lead to substantial site effects in multi-site studies \cite{Kruggel.2010, Wachinger.2019}. Clustering also occurs in biological data, such as when measurements are collected across different experimental batches \cite{Zaritsky.2021} or tissue samples \cite{Franks.2017}, and in environmental data collected across locations \cite{Gelman.2007}. 

If not properly handled in analysis, the cluster effects of non-\textit{iid} data can lead to erroneous conclusions. The so-called Simpson's paradox occurs when an association between two variables appears, disappears, or even reverses when analysis is performed at the population level vs. when analysis is stratified by cluster, indicating a confounding effect. This situation can lead to Type I (false positive) or Type II (false negative) findings in many situations, including clinical studies \cite{Holt.2016}, proteomics \cite{Franks.2017}, and economics \cite{Wagner.1982}. 

Despite these consequences, the machine learning community has generally ignored the problems underlying non-\textit{iid} data. Meanwhile, the traditional statistics community has addressed clustered data with \textit{mixed effects} models, which learn a combination of \textit{fixed} and \textit{random} effects. The most common of these is the linear mixed effects (LME) model, which builds upon the basic linear regression model. Suppose that we have data $X \in \mathbb{R}^{n \times p}$ with $n$ samples and $p$ independent variables (features), originating from $c$ clusters, and a dependent variable (target) $y \in \mathbb{R}^{n \times 1}$. We can define the following LME regression model; for a sample $i = 1, 2, ..., n$ originating from cluster $j = 1, 2, ..., c$ we have:
\begin{IEEEeqnarray}{rCl} 
\label{eq:lme1}
    \hat{y}_i &=& \beta_0 + x_{i,1} \beta_1 + ... + x_{i,p} \beta_p \IEEEnonumber\\
    &&+ u_{j, 0} + x_{i, 1} u_{j, 1} + ... + x_{i, p} u_{j, p} + \epsilon_i \\
    &=& \beta_0 + \boldsymbol{x}_i^\top \boldsymbol{\beta} + u_{j,0} + \boldsymbol{x}_i^\top \boldsymbol{u}_j + \epsilon_i \IEEEnonumber
\end{IEEEeqnarray}
where $\hat{y}_i$ is the predicted target, $\boldsymbol{x}_i^\top = [x_{i, 1}, ..., x_{i, p}]$ is the $p$-dimensional feature vector of the $i^{th}$ sample from $X$, and $\epsilon_i$ is the residual. The model contains two types of weights. The fixed effect intercept $\beta_0$ and slopes $\boldsymbol{\beta}^\top = [\beta_1, ..., \beta_p]$ are cluster-\textit{invariant} and apply globally to all samples. The random effects weights include the intercept $u_{j,0}$ and slopes $\boldsymbol{u}_j^\top = [u_{j, 1}, ..., u_{j, p}]$, whose values are \textit{specific} to each cluster $j$. The random effect weight values are assumed to follow a random distribution, most often a multivariate normal distribution with mean 0, i.e. $\boldsymbol{u} \sim N(0, \sigma)$. Consequently, the random effect weights $\boldsymbol{u}$ can be interpreted as cluster-specific offsets from the fixed effect weights $\boldsymbol{\beta}$. The LME model separates the variance explained by global associations from the inter-cluster variance, controls for correlated samples, and improves weight estimates \cite{Gelman.2007, Harrison.2018}. Unfortunately, proper handling of mixed effects in deep learning, delivering all of these gains, has gone unanswered. In this work, we describe how appropriate handling of mixed effects can address the inadequacies of deep learning models when applied to clustered data. 

\subsection{Related work}
Previous deep learning approaches for clustered data have key limitations. A naive but prevalent strategy is to insert cluster information into the model as an additional, one-hot encoded covariate \cite{Hancock.2020}. This increases data dimensionality, which may cause overfitting with a high number of clusters $c$ \cite{Hancock.2020, Simchoni.2021}, and it entangles the cluster-invariant and cluster-specific features within the model weights, hampering model interpretation. \textit{Domain adaptation} techniques train a model on a source domain (i.e. cluster), then adapt it in a subsequent training step to a target domain \cite{Wang.2018}. This yields an adapted model for each target domain but not a single unified model. It also does not scale easily to many domains or separate domain-invariant from domain-specific features, which also limits interpretability. \textit{Domain generalization} techniques address some of these weaknesses by producing a single generalized model agnostic to domain differences. Earlier approaches used gradient reversal layers, which modify backpropagation to maximize domain invariance \cite{Ganin.2016, Liu.12112018}. Other methods use \textit{meta-learning} to guide gradient descent in a direction that reduces the loss for all domains \cite{Li.2018, Liu.2020}.  However, these involve second-order optimization which vastly increase computational cost. A third category of domain generalization methods uses an adversarial classifier \cite{Tzeng.72120177262017, Kamnitsas.2017}. The adversarial classifier learns to classify domains from the latent features of the main model, while the main model learns features that maximize domain classification error. The common limitation of all domain generalization techniques is that they produce a model that has \textit{only} learned the domain-invariant features (fixed effects), while domain-specific information (random effects), are discarded. Our proposed framework captures this ignored information in a separate random effects subnetwork, while an adversarially-regularized subnetwork captures global fixed effects. We show that this adds predictive value and allows users to understand more about cluster variance in their data.

To date, there have been three prior approaches to incorporate mixed effects into deep learning. Xiong et al. proposed MeNet, a mixed effects convolutional neural network (CNN), for a gaze estimation dataset containing repeated images per subject \cite{Xiong.2019}. While improving accuracy, the method requires an expensive expectation-maximization algorithm with inversion of large covariance matrices ($n_j \times n_j$ where $n_j$ is the number of samples within each cluster). MeNet also only models random slopes and not intercepts. Next, Tran et al. proposed DeepGLMM, a mixed effects approach for dense feedforward neural networks (DFNNs) using Bayesian deep learning and variational inference for more efficient training \cite{Tran.2020}. Though theoretically capable of modeling both random slopes and intercepts, their applications only used models with random intercepts. Their experiments also lacked comparisons with other deep learning methods. Finally, Simchoni et al. proposed LMMNN, a mixed effects approach for both DFNNs and CNNs, and demonstrated a performance benefit across multiple applications. However, LMMNN is trained using expensive covariance matrix inversions, and their real-world applications use only random intercepts \cite{Simchoni.2021}.

There are several common limitations across the MeNet, DeepGLMM, and LMMNN approaches. These methods prioritize the improvement of predictive performance and ignore the additional interpretability afforded by mixed effects, such as quantification and visualization of inter-cluster variance. They also lack explicit guidance of the fixed effects to be cluster-invariant, so their resilience to confounded associations is unclear. Additionally, none of these works demonstrate models with both random slopes and intercepts or unsupervised learning models, such as autoencoders. Lastly, there are no specific recommendations for applying these models to new data that does not originate from the same clusters seen during training, which limits real-world utility where data from new clusters is frequently encountered.

\subsection{Contributions}
We propose an Adversarially-Regularized Mixed Effects Deep learning (ARMED) framework that generalizes across model archetypes and alleviates the shortcomings of the previous approaches. This framework contains three components that can be readily added to a conventional deep learning model with minimal modification of the existing architecture. First, inspired by domain generalization, we employ an adversarial classifier to regularize the model to learn cluster-invariant fixed effects. We show through simulations that this improves the separation of cluster-specific, potentially confounded features from cluster-invariant features. Second, we introduce a Bayesian random effects subnetwork to learn the cluster-specific features, and we demonstrate how it can quantify and visualize the variance across clusters. Third, we add another classifier which infers random effects for so-called ``unseen cluster" data, where samples originate outside the clusters seen during training. We demonstrate the advantages of our framework across 4 test cases using DFNNs, CNNs, and convolutional autoencoders, including simulations and three biomedical examples. In each case, we achieve not only the separation and identification of fixed and random effects, but also better predictive performance on data from seen clusters and better generalization to unseen clusters. 

\section{Methods}

\begin{figure}[!t]
    \centering
    \includegraphics[width=\columnwidth]{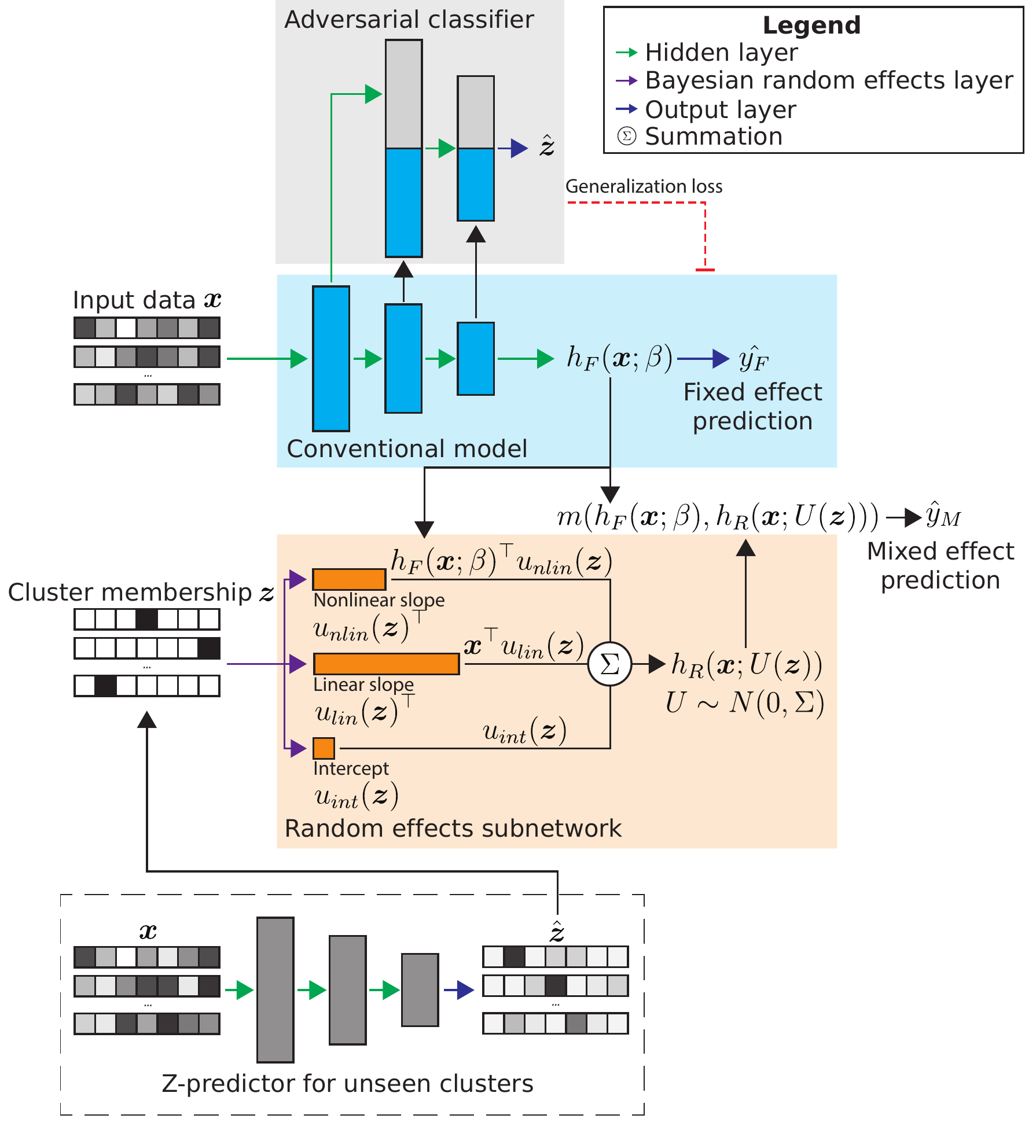}
    \caption{The ARMED framework for a generic neural network. The conventional model (blue area) predicts $\hat{y}_F$ from the data sample $\boldsymbol{x}$. Cluster membership of the sample is one-hot encoded into $\boldsymbol{z}$. The fixed effects subnetwork (blue $+$ gray areas) is constructed by adding an adversarial classifier (gray area) to predict cluster membership $\hat{\boldsymbol{z}}$. The original model is penalized through the generalization loss for learning features that allow cluster membership prediction. The random effects subnetwork (orange area) uses Bayesian layers to learn cluster-specific weights, dependent on $\boldsymbol{z}$, that follow zero-mean multivariate normal distributions. These weights can be formulated as nonlinear slopes multiplied by the fixed effects latent representation $h_F(X; \beta)$, linear slopes multiplied by $X$, and/or intercepts. The fixed and random effects are combined with a mixing function $m(...)$. For prediction on data from clusters unseen during training, $\boldsymbol{z}$ is inferred with a classifier (Z-predictor) trained on data from seen clusters.}
    \label{fig:general_framework}
\end{figure}

In general, a conventional feed-forward neural network computes a nonlinear transformation of the data $X$ through its layers (Fig. \ref{fig:general_framework}, blue area). We denote the output of the penultimate layer as $h(X, \beta) \in \mathbb{R}^{n \times q}$, where $\beta$ contains all learned weights up to and including this layer and $q$ is the number of neurons. For a typical regression or classification task, a final linear or softmax output layer $o$ then transforms $h(X, \beta)$ into the final prediction output $\hat{\boldsymbol{y}}$:
\begin{equation*} \label{eq:nn_output}
    \hat{\boldsymbol{y}} = o(h(X, \beta))
\end{equation*}
During training, the model finds the weights, $\beta$, which minimize a given loss function quantifying the \textbf{e}rror  for the predictive task, $\mathcal{L}_e(\boldsymbol{y}, \hat{\boldsymbol{y}})$.

To encode cluster membership information for a dataset with $n$ samples and $c$ clusters, we introduce a one-hot encoded design matrix $Z \in \mathbb{R}^{n \times c}$, where $Z_{i,j} = 1$ if sample $i$ belongs to cluster $j$ and $Z_{i,j} = 0$ otherwise. The following sections present a description of the ARMED framework components, agnostic to model architecture. These components include the fixed effects subnetwork $h_F$, including a conventional neural network and an adversarial classifier $a$ that together learn cluster-invariant features, the random effects subnetwork $h_R$ for learning $Z$-dependent cluster-specific features, the mixing function $m$ that combines the fixed and random effects for prediction, and the Z-predictor used to apply random effects to new clusters. 

\subsection{Fixed effects subnetwork}
First, we add an adversarial classifier (Fig. \ref{fig:general_framework}, gray area) to the conventional model (Fig. \ref{fig:general_framework}, blue area) to enforce the learning of cluster-invariant fixed effects, creating the \textit{F}ixed effects subnetwork $h_F(X; \beta)$. This is based on the adversarial learning technique for domain generalization \cite{Tzeng.72120177262017, Kamnitsas.2017}. For a neural network with $L$ layers, let 
\begin{equation*} \label{eq:nn}
    H_F(X; \beta_F) = [h_{F, 1}(X; \beta_{F, 1}), ..., h_{F, L}(X; \beta_{F, L})]
\end{equation*}
represent the collected outputs of each layer, where $\beta_{F, l}$ contains the weights up to the $l$th layer. We define an adversarial classifier $a$ which predicts a sample's cluster membership from these layer outputs, $\hat{Z} = a(H_F(X; \beta_F); \beta_A)$, where $\beta_A$ contains the weights for this adversary. The adversary is trained to minimize the categorical cross-entropy loss:
\begin{IEEEeqnarray}{rCl}
    &\mathcal{L}_{CCE}&(Z, \hat{Z}) \IEEEnonumber\\
    &{=}& - \frac{1}{n} \sum_{i = 1}^n \sum_{j=1}^c Z_{i,j} \log (\hat{Z}_{i, j}) + (1 - Z_{i, j}) \log (1 - \hat{Z}_{i, j}) \IEEEnonumber
\end{IEEEeqnarray}
Meanwhile, the main model is penalized for learning features that allow the adversary to predict cluster membership. It must \textit{maximize} this cross-entropy, which we call the cluster generalization loss. The resulting training objective of the fixed effects subnetwork is
\begin{equation} \label{eq:fe_loss}
    \mathcal{L}_e(\boldsymbol{y}, \hat{\boldsymbol{y}}_F) - \lambda_g \mathcal{L}_{CCE}(Z, \hat{Z})
\end{equation}
where the hyperparameter $\lambda_g$ controls the weight of the generalization loss. We use $\hat{\boldsymbol{y}}_F$ to denote the prediction output of this fixed effects subnetwork. 

\subsection{Random effects subnetwork} \label{section:re_subnet}

We next define a second subnetwork to learn the \textit{R}andom effects, $h_R(X; U(Z))$ with cluster-specific weights $U(Z)$ (Fig. \ref{fig:general_framework}, orange area). The cluster-specific values for each individual weight $u(Z)$ in $U(Z)$ are assumed to follow a normal distribution with mean 0, i.e. $u(Z) \sim N(0, \sigma)$ where $\sigma$ represents the inter-cluster variance of each weight. Collectively, $\Sigma$ contains the inter-cluster variance for all weights in $U(Z)$. We implement these weights using a Bayesian formulation. We specify a zero-mean normal prior distribution for each weight $p(U) \sim N(0, \sigma_p)$ with the fixed prior variance $\sigma_p$ as a global hyperparameter. The posterior distribution $p(U | X)$ is then learned through variational inference, which reframes Bayesian modeling as an optimization problem that can be efficiently handled through gradient descent \cite{Kingma.12202013, Blei.2017}. The objective of variational inference is to learn a surrogate posterior $q(U)$, here a multivariate normal distribution, which closely approximates the true posterior $p(U | X)$, where ``closeness" is measured by the Kullback-Leibler (KL) divergence: 
\begin{equation*} \label{eq:kld}
    D_{\text{KL}}(q(U) || p(U | X)) = \int q(U) \log \frac{q(U)}{p(U | X)} dU
\end{equation*}
Minimizing $D_{\text{KL}}(q(U) || p(U | X))$ directly is impossible because computing the posterior through Bayes Rule, $p(U | X) = \frac{p(X | U) p(U)}{p(X)}$, involves the intractable marginalization $p(X)$. Instead, variational inference maximizes the Evidence Lower Bound (ELBO) which contains fully tractable and differentiable quantities:
\begin{equation*} \label{eq:elbo}
    \text{ELBO} = \mathbb{E}_q [\log p(X | U)] - D_{\text{KL}}(q(U) || p(U))
\end{equation*}
where the first right-hand term is the log-likelihood and the second term is the KL divergence between the surrogate posterior and the prior. For gradient descent, we minimize the negative ELBO and let $\mathcal{L}_e(\boldsymbol{y}, \hat{\boldsymbol{y}})$ represent the first term, i.e. the negative log-likelihood loss. This yields the following objective:
\begin{equation} \label{eq:re_loss}
    \mathcal{L}_e(\boldsymbol{y}, \hat{\boldsymbol{y}}) + \lambda_{K} D_{\text{KL}}(q(U) || p(U))
\end{equation}
with the hyperparameter $\lambda_K$ controlling the strength of the regularization to the prior. Note that our method based on variational inference does not require expensive inversions of covariance matrices as in MeNet and LMMNN \cite{Xiong.2019, Simchoni.2021}.

The architecture of this subnetwork will depend on the types of random effects to be modeled. Nonlinear random effects slopes can be modeled as weights multiplied by the fixed effects latent representation $h_F(X; \beta)$:
\begin{equation} \label{eq:re_nonlinear}
    h_{R, nlin}(\boldsymbol{x}_i; u_{nlin}(\boldsymbol{z}_i)) = h_F(\boldsymbol{x}_i; \beta)^\top u_{nlin}(\boldsymbol{z}_i)
\end{equation}
where $\boldsymbol{z}_i$ and $\boldsymbol{x}_i$ are the rows in $Z$ and $X$ for the $i$th sample and $u_{nlin}(\boldsymbol{z}_i) \in \mathbb{R}^{q \times 1}$ returns the slopes for cluster $\boldsymbol{z}_i$, $q$ being the number of output neurons of $h_F(\boldsymbol{x}_i; \beta)$. A random intercept is modeled simply as a weight:
\begin{equation} \label{eq:re_intercept}
    h_{R, int}(u(\boldsymbol{z}_i)) = u_{int}(\boldsymbol{z}_i)
\end{equation}
where $u_{int}(\boldsymbol{z}_i)$ is a scalar value. Additionally, for tabular data, we can model linear random effects slopes multiplied directly with $X$, which allows each slope to be interpreted directly with respect to a corresponding input variable:
\begin{equation} \label{eq:re_linear}
    h_{R, lin}(\boldsymbol{x}_i; u_{lin}(\boldsymbol{z}_i)) = \boldsymbol{x}_i^\top u_{lin}(\boldsymbol{z}_i)
\end{equation}
where $u_{lin}(\boldsymbol{z}_i) \in \mathbb{R}^{p \times 1}$ returns the slopes for cluster $\boldsymbol{z}_i$. The random effects subnetwork outputs the sum of these random effects: 
\begin{IEEEeqnarray}{rCl}
    h_R(\boldsymbol{x}_i; U(\boldsymbol{z}_i)) &{}={}& h_{R, nlin}(\boldsymbol{x}_i; u_{nlin}(\boldsymbol{z}_i)) \IEEEnonumber\\
    &&{+} h_{R, lin}(\boldsymbol{x}_i; {u}_{lin}(\boldsymbol{z}_i)) \\
    &&{+} h_{R, int}(u_{int}(\boldsymbol{z}_i)) \IEEEnonumber
\end{IEEEeqnarray}
These three cases will apply to most models with a dense penultimate layer producing a vector-form $h_F(X; \beta)$. For models such as autoencoders, we describe in the Supplemental Materials how random effects can be readily applied across multiple convolutional layers (Section 3.1.3, Fig. S4). 

\subsection{Combining fixed and random effects}

We construct the final ARMED model by combining the outputs of the fixed effects and random effects subnetworks. In the linear model of Eq. \ref{eq:lme1}, random and fixed effects were combined through addition. For greater flexibility here, we substitute the addition in Eq. \ref{eq:lme1} with a more general mixing function $m(...)$.
\begin{equation} \label{eq:ARMED1}
    \hat{\boldsymbol{y}}_{M} = m(h_F(X; \beta), h_R(X; U(Z)))
\end{equation}
For example, in the following binary classification applications, we use a nonlinear analog of Eq. \ref{eq:lme1}. We add $h_R(X; U(Z))$ to the logit of $\hat{\boldsymbol{y}}_F$ (equal to $h_F(\boldsymbol{x}_i; \beta)^\top \beta_L$ where $\beta_L$ are the weights of the output layer), then apply the sigmoid activation function:
\begin{equation*}
    \hat{\boldsymbol{y}}_{M} = sigmoid \left(h_F(\boldsymbol{x}_i; \beta)^\top \beta_L + h_R(\boldsymbol{x}_i; U(\boldsymbol{z}_i)) \right)
\end{equation*}
The objective function is obtained by combining Eq. \ref{eq:fe_loss} and Eq. \ref{eq:re_loss}:
\begin{IEEEeqnarray}{rCl} \label{eq:me_loss}
    &&\mathcal{L}_e(\boldsymbol{y}, \hat{\boldsymbol{y}}_M) + \lambda_F \mathcal{L}_e(\boldsymbol{y}, \hat{\boldsymbol{y}}_F) \\
    &&{-} \lambda_g \mathcal{L}_{CCE}(Z, \hat{Z}) + \lambda_{K} D_{\text{KL}}(q(U) || p(U)) \IEEEnonumber
\end{IEEEeqnarray}
The second term ensures that the fixed effect subnetwork will still be capable of prediction on its own so that the fixed effect features will be meaningful in later analyses. The loss weight $\lambda_F < 1$ balances the fixed effect error with the mixed effect error $\mathcal{L}_e(\boldsymbol{y}, \hat{\boldsymbol{y}}_M)$. 

ARMED includes these hyperparameters: the generalization loss weight $\lambda_g$, the KL divergence weight $\lambda_K$, the fixed effect prediction error weight $\lambda_F$, and the prior distribution variance $\sigma_p$. Usage of linear vs. nonlinear slopes must also be considered. In practice, we find that these can be easily tuned for model performance using standard hyperparameter optimization approaches, such as random search or Bayesian optimization, and appropriate cross-validation.

\subsection{Prediction on unseen clusters}
The previous mixed effects deep learning approaches provide no method for using the learned random effects when predicting on data not from clusters seen during training, i.e. not included in $Z$ \cite{Xiong.2019, Tran.2020, Simchoni.2021}. The authors of LMMNN propose to use only the fixed effects of their model on unseen clusters \cite{Simchoni.2021}. While the learned fixed effects, by definition, represent population-average associations, new data is not necessarily free of cluster effects and performance may be improved by fully utilizing the learned random effects. We propose to infer $Z$ for unseen cluster data using a classifier we call the \textit{Z-predictor}. We train this classifier to predict $Z$ from $X$ on the data from seen clusters, then use it to infer $Z$ for data from unseen clusters. The unthresholded softmax predictions from the classifier provide a weighted combination of seen clusters that are most similar to each unseen cluster sample. In our applications, the Z-predictor uses the same architecture as the adversarial classifier.

\subsection{Applications}

\subsubsection{Applications of ARMED to dense feedforward neural networks}
Our first architectural application of ARMED is to a dense feedforward neural network (DFNN), which is suited to tabular data such as clinical measurements or pre-engineered image features. We describe the specifics of the ARMED-DFNN architecture in Fig. S1 and the Supplemental Materials 3.1.1. \par \vspace{5pt}
\noindent \textbf{Spiral classification simulations:}
First, we evaluated the ARMED-DFNN on a simulated classification problem where cluster effects can be controlled, model-learned information can be compared to ground truth, and known confounded features can be added. The simulations are built upon the well-known spiral classification problem, where points must be classified into one of two spirals based on their coordinates $x_1$ and $x_2$ \cite{Lang.1988}. We simulated a nonlinear random effect by dividing the points into 10 clusters and randomly varied the spiral radius across clusters (Fig. S2). There were 3 variations of this simulation: 1) spiral radii varied across clusters, 2) spiral radii varied across clusters and spiral labels were inverted in half of the clusters (a more severe random effect), and 3) spiral radii varied across clusters and 2 known confounded probe features $x_3$ and $x_4$ were added. These probes created a spurious association between cluster and label but were not associated with the underlying spiral functions. Further details on these simulations can be found in the Supplemental Materials 3.2. Because we have defined the random effects to be nonlinear, we used an ARMED-DFNN architecture with \textit{nonlinear} random slopes (Eq. \ref{eq:re_nonlinear}) and a random intercept (Eq. \ref{eq:re_intercept}). 

To test the ability of the fixed effects subnetwork to correctly downweight these confounded probes, we measured feature importance by computing the gradient of the model output with respect to the input features \cite{Dimopoulos.1995, Olden.2004}. Features with larger gradient magnitudes are more important in forming the model output. We compared the importance of each confounded probe ($x_3$ and $x_4$) to that of the least important true feature ($x_1$ or $x_2$). 
\par \vspace{5pt}
\noindent \textbf{Mild cognitive impairment conversion prediction:}
For a complementary real-world application, the ARMED-DFNN was used to predict the future development of full Alzheimer's Disease (AD) in subjects with mild cognitive impairment (MCI). MCI is an early stage of cognitive decline that may progress to dementia. Our target was to distinguish progressive MCI (pMCI), where a subject converts to AD within 24 months of baseline observation, from stable MCI (sMCI), where the subject does not convert within 24 months. We used data from the Alzheimer's Disease Neuroimaging Initiative, which includes baseline demographic information, cognitive scores, neuroimaging measurements, and biomarker measurements, as well as longitudinal diagnoses for each participant, acquired with informed consent and institutional review board approval (Supplemental Materials 3.3.1). The training dataset came from the largest 20 study sites, and we used site as the random effect cluster. Inter-site variance has been shown to affect cognitive scores, which are sensitive to judgments by human raters, and neuroimaging, which is sensitive to MRI scanner parameters \cite{Connor.2008, Schafer.2011, ThibeauSutre.2022}.  We held out the remaining 34 sites to evaluate model performance on sites unseen during training. Performance metrics included area under the receiver operating characteristic curve (AUROC), balanced accuracy, sensitivity, and specificity. For this application, we used an architecture with \textit{linear} random slopes (Eq. \ref{eq:re_linear}) and a random intercept (Eq. \ref{eq:re_intercept}). These were chosen to allow direct interpretation of the learned random slopes and inter-site variance for each input feature. 

As with the spiral simulations, we subsequently added simulated confounded probe features to test how well each model could downweight known confounded features. We generated 5 confounded probes that were nonlinearly associated with site and with the probability of being labeled pMCI but had no real biological relevance (Supplemental Materials 3.3.1). We then compared how highly each model ranked the probes based on feature importance (gradient magnitudes). 

\subsubsection{Application of ARMED to convolutional neural networks}
We next applied our approach to a convolutional neural network (CNN), another important deep learning archetype, creating an ARMED-CNN capable of learning nonlinear random slopes and random intercepts. Architecture details are described in Fig. 3 and Supplemental Materials 3.1.2. 

We applied the ARMED-CNN to the classification of AD vs. cognitively normal (CN) structural MRI, with study site as the random effect cluster. We acquired T1-weighted MRI from 12 sites in the ADNI dataset (inclusion criteria and preprocessing details are in Supplemental Materials 3.3.2). These 12 sites were selected to emphasize the confounding site effect, where sites using General Electric MRI scanners had a greater proportion of AD subjects compared to sites using Philips or Siemens scanners (Table S1). The remaining 51 sites were held out to evaluate performance on sites unseen during training. We extracted a two-dimensional coronal slice through the hippocampi from each image. Performance metrics included AUROC, balanced accuracy, sensitivity, and specificity. 

\subsubsection{Application of ARMED to autoencoders}
To demonstrate our framework on unsupervised learning models, we developed a mixed effects autoencoder. Our fourth application was the melanoma live-cell image compression and phenotypic classification problem described in \cite{Zaritsky.2021}. In this work, the authors used a convolutional autoencoder to compress the images into a vector latent representation, then trained a classifier to label cells as having either high or low metastatic efficiency. They revealed that batch effects are prominent in this dataset, due to discrepancies between image batches acquired across different days, and that the latent representations strongly segregated by batch. The dataset is described further in Supplemental Materials 3.4. The training data from the melanoma cell image dataset contained images acquired over 13 days (batches), and the remaining 11 days were held out as unseen batches. 

We extended their autoencoder architecture by connecting the metastatic efficiency classifier directly to the autoencoder and training the autoencoder-classifier (AEC) end-to-end. We then applied our ARMED framework to create an ARMED-AEC, containing a fixed effects subnetwork that produces batch-invariant latent representations and a random effects subnetwork that learns how the batch effects alter image appearance (Fig. S4). Our hypothesis was that the modeling of mixed effects would improve classification performance over the base AEC. This architecture is described in Supplemental Materials 3.1.3.  

In addition to evaluating the reconstruction error (MSE) and phenotype prediction performance (AUROC), we also measured how strongly each model's latent representations clustered by batch. We computed the Davies-Bouldin (DB) score, where lower values indicate stronger clustering \cite{Davies.1979},  and the Calinksi-Harabasz (CH) score, where higher values indicate stronger clustering  \cite{Calinski.1974}. Consequently, we desire a higher DB score and lower CH score to achieve batch-invariant latent representations. 

\subsection{Compared methods and ablation tests}
In each application, we compared the proposed mixed effects model with the following approaches. First, we tested a conventional neural network where the cluster membership $Z$ is disregarded and data is assumed to be \textit{iid}. Second, for the DFNN and CNN, we tried the ``cluster input" approach of treating the one-hot cluster membership $Z$ as a categorical covariate, i.e. an additional model input. For the DFNN, $Z$ was simply concatenated to $X$. For the CNN, $Z$ was concatenated to flattened output of the last convolutional layer, before the dense hidden layer. When evaluating on unseen clusters, we used the inferred $Z$ from the Z-predictor. Third, we also compared to meta-learning domain generalization (MLDG) \cite{Li.2018}. However, due to the high computational cost of second-order gradients in MLDG (training took 10 times longer than the conventional DFNN) and poor performance, we dropped the MLDG comparison for the other applications, after the spiral simulation application. Fourth, we tested a domain adversarial (DA) neural network, i.e. the fixed effect subnetwork by itself. Despite regularization to learn only fixed effects, it does not model any cluster-specific random effects. Finally, for the DFNN and CNN, we tested MeNet \cite{Xiong.2019} and LMMNN \cite{Simchoni.2021}. For the autoencoder, only the proposed ARMED approach has a suitable adaptation. 

Additionally, we performed two ablation tests of the proposed mixed effects approach. We first trained the ARMED models without the adversarial classifier (``w/o Adv.") to test the necessity of the generalization loss to learn fixed effects. Additionally, we evaluated the model using randomly-assigned cluster memberships in $Z$ (``randomized Z"). For data from seen clusters, this tested whether the model truly learned cluster-specific effects. For data from unseen clusters, this tested the impact of using the Z-predictor to infer cluster membership. 

\section{Results}

\begin{table*}[!t]
    \renewcommand{\arraystretch}{1.25}
    \centering
    \begin{threeparttable}
    \caption{Spiral simulation results with 10-fold cross-validation. The best results in each simulation are \textbf{bolded}.}
    \label{table:spiral_results}
    \begin{tabular}{@{\extracolsep{5pt}}lororor@{}}
    \hline
                      & \multicolumn{2}{l}{\begin{tabular}[c]{@{}l@{}}Simulation 1:\\cluster-specific radii\end{tabular}} 
                      & \multicolumn{2}{l}{\begin{tabular}[c]{@{}l@{}}Simulation 2:\\cluster-specific radii with inversions\end{tabular}} 
                      & \multicolumn{2}{l}{\begin{tabular}[c]{@{}l@{}}Simulation 3:\\Simulation 1 + confounded features\end{tabular}} \\
    \cline{2-3} \cline{4-5} \cline{6-7}
    Model                   & \dheader{Mean acc. (\%)} & 95\% CI & \dheader{Mean acc. (\%)} & 95\% CI & \dheader{Mean acc. (\%)} & 95\% CI \\ 
    \hline
    Conventional DFNN       & 76.4          & 75.8 - 76.9   & 52.3         & 51.3 - 53.2 & 72.1          & 70.2 - 74.1 \\
    Cluster input DFNN      & 67.6          & 65.8 - 69.3   & 67.1         & 66.1 - 68.2 & \boldo{76.4}  & 75.6 - 77.3 \\
    MLDG                    & 62.8          & 60.9 - 64.7   & 50.2         & 49.9 - 50.5 & 65.8          & 64.9 - 66.7 \\
    DA-DFNN                 & 75.3          & 72.5 - 78.0   & 49.5         & 48.8 - 50.2 & 66.3          & 63.6 - 69.0 \\
    MeNet                   & 77.4          & 76.8 - 78.0   & 53.3         & 52.1 - 54.5 & 73.0          & 71.0 - 75.0 \\
    LMMNN                   & 50.0          & 50.0 - 50.0   & 50.0         & 50.0 - 50.0 & 47.0          & 46.7 - 47.3 \\
    ARMED-DFNN              & 78.8          & 78.0 - 79.6   & 65.0         & 61.2 - 68.8 & 74.5          & 72.2 - 76.9 \\
    \hspace{1em} w/o Adv.   & \boldo{79.3}  & 76.9 - 81.8   & \boldo{69.9} & 67.3 - 72.5 & 68.5          & 67.5 - 69.4  \\
    \hspace{1em} randomized Z & 76.7        & 75.6 - 77.8   & 50.6         & 49.7 - 51.6 & 69.7          & 66.7 - 72.8 \\ 
    \hline
    \end{tabular}
    \begin{tablenotes}
        \footnotesize
        \item DFNN: dense feedforward neural network; MLDG: meta-learning domain generalization; DA: domain adversarial; Adv.: adversary; acc.: accuracy; CI: confidence interval
    \end{tablenotes}
    \end{threeparttable}
\end{table*}

\begin{table}[t]
    \renewcommand{\arraystretch}{1.25}
    \centering
    \caption{Sensitivity to confounded probe features in spiral simulation 3. Paired T-tests were computed to compare the feature importance of each confounded probe with the least important true feature. Positive and larger T-statistics are desired, indicating the model placed higher importance on the true feature than the confounded probe.}
    \label{table:spiral_sim_3}
    \begin{tabular}{@{}l t r t r@{}}
    \hline
                        & \multicolumn{2}{c}{Probe $x_3$} & \multicolumn{2}{c}{Probe $x_4$} \\
    Model               & \dheader{T-statistic} & $p$-value & \dheader{T-statistic} & $p$-value \\
    \hline
    Conventional DFNN       & -0.232         & 0.822           & -0.861            & 0.411           \\
    Cluster input DFNN      & 5.346          & \textless 0.001 & 5.042             & \textless 0.001 \\
    MLDG                    & -16.573        & \textless 0.001 & -14.535           & \textless 0.001 \\
    DA-DFNN                 & 4.072          & 0.003           & 3.832             & 0.004           \\
    MeNet                   & 7.923          & \textless 0.001 & 4.541             & \textless 0.001 \\
    LMMNN                   & 8.369          & \textless 0.001 & 5.090             & \textless 0.001 \\
    ARMED-DFNN              & \boldt{12.632} & \textless 0.001 & \boldt{18.173}    & \textless 0.001 \\ 
    \hspace{1em} w/o Adv.   & 1.390          & 0.198           & 0.343             & 0.740           \\
    \hline
    \end{tabular}
\end{table}

\subsection{Spiral classification simulations}
\begin{figure*}[ht]
    \centering
    \includegraphics[width=0.9\textwidth]{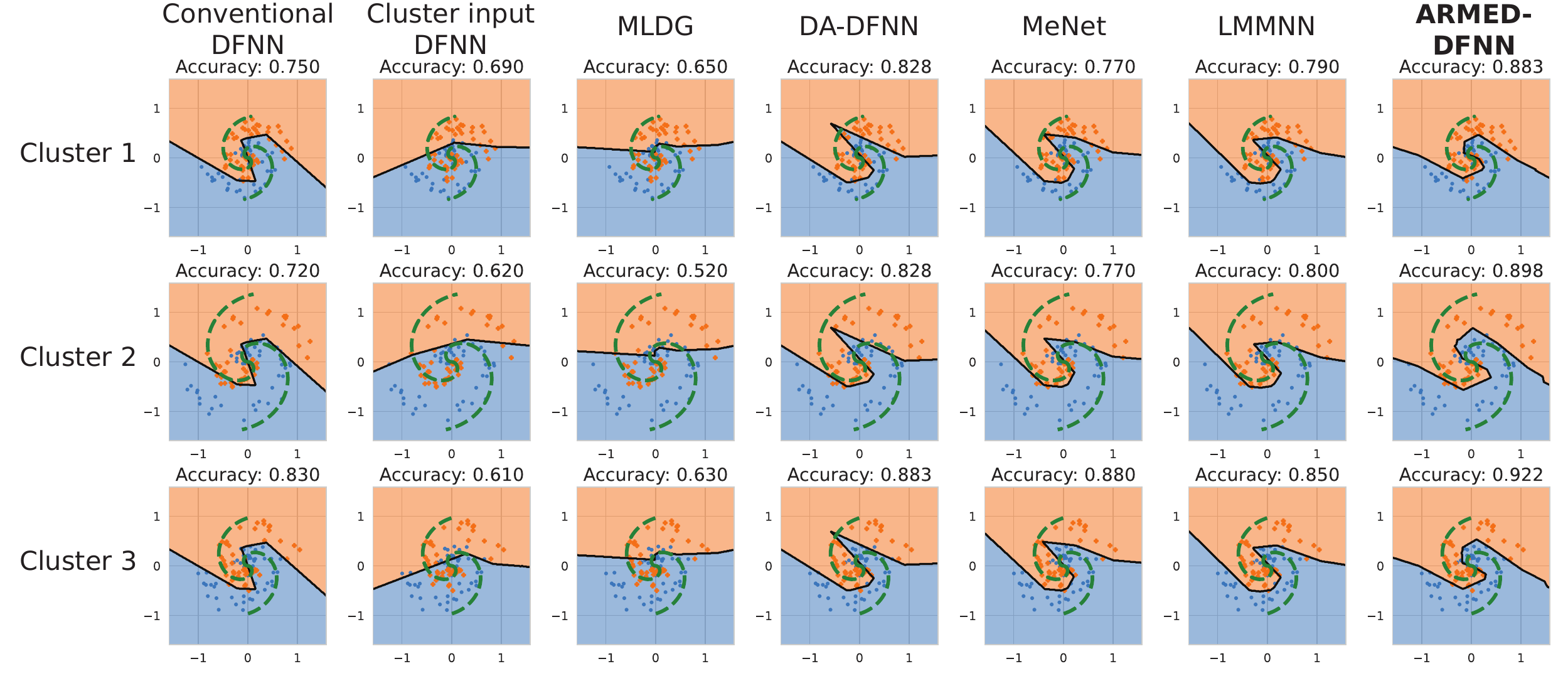}
    \caption{Decision boundaries learned by each model in spiral simulation 1, where spiral radii varies across clusters as a random effect. Each row illustrates one of 3 representative clusters from 10 total simulated clusters. Each column contains the decision boundaries (black solid line) learned by one model. The green dashed line illustrates the true decision boundary, computed as the midpoint between the two spirals. Only the ARMED-DFNN was able to learn the appropriate cluster-specific decision boundaries.}
    \label{fig:spiral_sim_1}
\end{figure*}

The classification accuracy of each model, with 10-fold cross-validation, is presented in Table \ref{table:spiral_results}. In simulation 1 (random cluster-specific radii distributed around 1), the ARMED-DFNN outperformed all other models and had statistically significantly higher accuracy than the second-best model, MeNet (78.8\% vs. 77.4\%, $p = 0.003$ in paired T-test). It was also uniquely able to learn appropriate cluster-specific decision boundaries that scaled in size with the cluster-specific spiral radii (Fig. \ref{fig:spiral_sim_1}). For example, cluster 1 (left column) has the smallest ground truth radius (green dashed line), and while cluster 2 (middle column) has the largest true radius, and the ARMED-DFNN uniquely learned this difference.  In simulation 2 (greater inter-cluster variance, spiral labels inverted in half), only the cluster input DFNN, MeNet, and ARMED-DFNN achieved accuracy substantially higher than chance (50\%), with 67.1\%, 53.3\%, and 65.0\% respectively. The cluster input DFNN and ARMED-DFNN statistically significantly outperformed MeNet ($p \ll 0.001$), but did not differ significantly from each other at $p < 0.05$. In simulation 3 (confounded probe features added), the cluster input ranked first (76.4\%), followed by the ARMED-DFNN (74.5\%) and MeNet (73.0\%). However, the ARMED-DFNN more effectively downweighted the 2 confounded features compared to the true features (T-statistic = 12.631 and 18.173) compared to the cluster input DFNN (T-statistic = 5.346 and 5.042) and MeNet (T-statistic = 7.923 and 4.541) (Table \ref{table:spiral_sim_3}). The conventional and meta-learning models placed greater importance on the confounded than the true features. 

In ablation tests, removing adversarial regularization non-significantly improved the accuracy of the ARMED-DFNN in simulations 1 and 2, but decreased accuracy in simulation 3. It also worsened the separation of confounded and true features in simulation 3 (T = 1.390 and 0.343). Using randomly assigned cluster memberships in $Z$ uniformly decreased performance, confirming that the ARMED-DFNN learned necessary cluster-specific information.  

\subsection{MCI conversion prediction}

\begin{table*}[t]
    \renewcommand{\arraystretch}{1.25}
    \centering
    \begin{threeparttable}
    \caption{Prediction of stable vs. progressive mild cognitive impairment. Confidence intervals were computed through 10$\times$10-fold nested cross-validation. Sensitivity and specificity were computed at the Youden point. The best results for each metric are \textbf{bolded}.}
    \label{table:mci}
    \begin{tabular}{@{\extracolsep{5pt}}ltrororor@{}}
    \hline
    & \multicolumn{2}{c}{AUROC} & \multicolumn{2}{c}{Balanced accuracy (\%)} & \multicolumn{2}{c}{Sensitivity (\%)} & \multicolumn{2}{c}{Specificity (\%)} \\
    \cline{2-3} \cline{4-5} \cline{6-7} \cline{8-9}
    Model                       & \dheader{Mean} & 95\% CI & \dheader{Mean} & 95\% CI & \dheader{Mean} & 95\% CI & \dheader{Mean} & 95\% CI \\ \hline
    \multicolumn{9}{c}{Seen sites} \\
    Conventional DFNN           & 0.884         & 0.836 - 0.931   & 80.8         & 74.6 - 87.0   & \boldo{81.2} & 68.3 - 94.1   & 80.3          & 74.7 - 86.0 \\
    Cluster input DFNN          & 0.866         & 0.819 - 0.914   & 81.3         & 75.8 - 86.8   & 80.2         & 68.6 - 91.7   & 82.4          & 77.5 - 87.3 \\
    DA-DFNN                     & 0.811         & 0.745 - 0.876   & 75.5         & 68.9 - 82.2   & 74.9         & 62.3 - 87.6   & 76.1          & 69.0 - 83.2 \\
    MeNet                       & 0.830         & 0.780 - 0.880   & 75.5         & 68.3 - 82.7   & 73.7         & 59.0 - 88.4   & 77.3          & 71.7 - 82.9 \\
    LMMNN                       & 0.860         & 0.824 - 0.896   & 79.4         & 72.2 - 86.6   & 73.9         & 59.7 - 88.1   & 84.9          & 81.6 - 88.1 \\
    ARMED-DFNN                  & \boldt{0.926} & 0.901 - 0.951   & \boldo{81.9} & 77.7 - 86.1   & 76.5         & 67.6 - 85.3   & \boldo{87.4}  & 84.5 - 90.2 \\ 
    \hspace{1em} w/o Adv.       & 0.919         & 0.891 - 0.946   & 81.4         & 76.8 - 86.1   & 74.5         & 64.6 - 84.4   & \boldo{88.4}  & 85.4 - 91.4 \\
    \hspace{1em} randomized Z   & 0.889         & 0.862 - 0.916   & 79.1         & 73.9 - 84.2   & 73.9         & 64.0 - 83.9   & 84.2          & 80.2 - 88.2 \\ 
    \hline
    \multicolumn{9}{c}{Unseen sites} \\
    Conventional DFNN           & 0.806         & 0.786 - 0.825   & 73.9         & 71.9 - 76.0   & \boldo{76.2} & 73.4 - 78.9   & 71.7          & 68.5 - 74.8 \\
    Cluster input DFNN          & 0.796         & 0.776 - 0.816   & 74.4         & 72.7 - 76.2   & 75.4         & 72.5 - 78.4   & 73.4          & 71.6 - 75.2 \\
    DA-DFNN                     & 0.723         & 0.665 - 0.780   & 67.9         & 63.2 - 72.6   & 64.7         & 52.7 - 76.8   & 71.1          & 67.4 - 74.7 \\
    MeNet                       & 0.750         & 0.693 - 0.807   & 70.2         & 65.6 - 74.9   & 66.0         & 57.7 - 74.4   & 74.5          & 69.8 - 79.1 \\
    LMMNN                       & 0.811         & 0.805 - 0.817   & 74.6         & 73.6 - 75.7   & 71.1         & 68.1 - 74.2   & 78.1          & 76.9 - 79.3 \\
    ARMED-DFNN                  & \boldt{0.837} & 0.833 - 0.842   & \boldo{75.6} & 74.1 - 77.1   & 72.4         & 67.6 - 77.1   & 78.8          & 76.6 - 80.9 \\ 
    \hspace{1em} w/o Adv.       & \boldt{0.838} & 0.827 - 0.848   & 73.5         & 72.5 - 74.5   & 65.4         & 62.9 - 67.8   & \boldo{81.7}  & 80.7 - 83.3 \\
    \hspace{1em} randomized Z   & 0.830         & 0.822 - 0.837   & 74.6         & 73.3 - 75.9   & 69.8         & 65.0 - 74.5   & 79.5          & 77.0 - 82.0 \\
    \hline
    \end{tabular}
    \begin{tablenotes}
        \footnotesize
        \item DFNN: dense feedforward neural network; MLDG: meta-learning domain generalization; DA: domain adversarial; Adv.: adversary; AUROC: area under receiver operating characteristic curve; CI: confidence interval. 
    \end{tablenotes}
    \end{threeparttable}
\end{table*}

The performance of each model in classifying pMCI vs. sMCI, over 10$\times$10 nested cross-validation folds, is compared in Table \ref{table:mci}. On study sites \textit{seen} during training, the ARMED-DFNN outperformed all other models in AUROC, accuracy, and specificity (Table \ref{table:mci}, top). The AUROC of the ARMED-DFNN was statistically significantly higher than that of the second-best model, the conventional DFNN (0.926 vs. 0.884, $p = 0.048$). On held-out study sites \textit{unseen} during training, the ARMED-DFNN again outperformed all other models in AUROC, accuracy, and specificity (Table \ref{table:mci}, bottom). The AUROC of the ARMED-DFNN was statistically significantly higher than that of the second-best conventional DFNN (0.837 vs. 0.806, $p = 0.007$). The DA-DFNN performed the poorest on both seen and unseen sites, with AUROC of 0.811 and 0.723 respectively. 

Removing the adversarial regularization of the ARMED-DFNN reduced AUROC (0.926 to 0.919) and accuracy (81.9\% to 81.4\%) on seen sites and accuracy (75.6\% to 73.5\%) and sensitivity (72.4\% to 65.4\%) on unseen sites. On seen sites, randomizing the site assignments reduced all metrics, including AUROC from 0.926 to 0.889. On unseen sites, using random instead of inferred site assignments also reduced all metrics including sensitivity from 72.4\% to 69.8\%. 

\begin{figure}[h]
    \centering
    \includegraphics[width=\columnwidth]{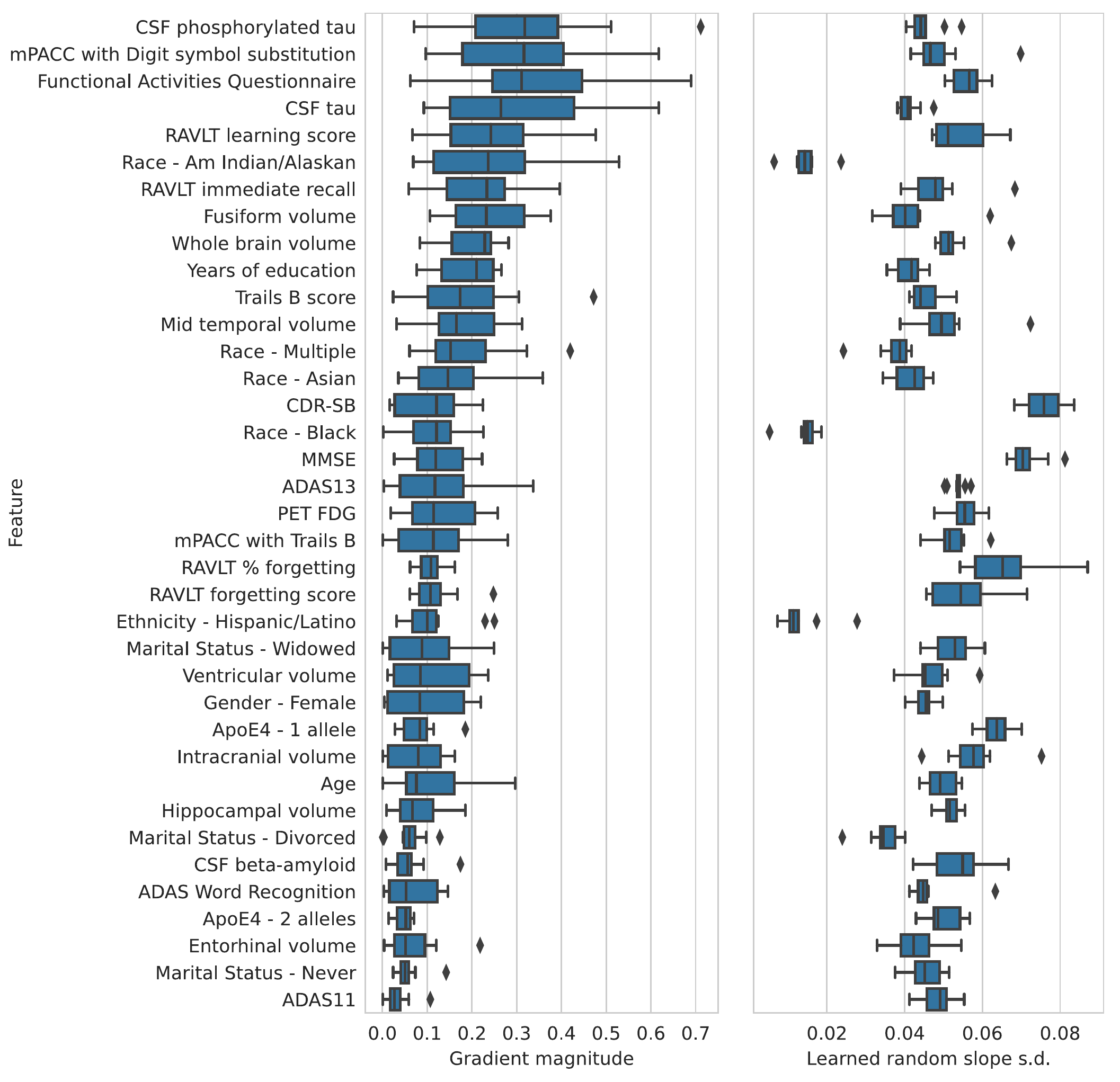}
    \caption{Feature importance and random slope variance for the ARMED-DFNN predictor of stable vs. progressive mild cognitive impairment. \textit{left}) Features are ranked by descending median feature importance (gradient magnitude) across 10 cross-validation folds, measured from the fixed effects subnetwork. \textit{right}) The inter-site variance of each feature's random slopes. See Supplemental section 3.3.1 for abbreviations.}
    \label{fig:mci_me_features}
\end{figure}

We examined the feature importance ranking, based on the fixed effects subnetwork, and learned site-specific random slopes, based on the random effects subnetwork, of the ARMED-DFNN (Fig. \ref{fig:mci_me_features}). Demographic features including, race, ethnicity, and marital status had especially low inter-site variance. Cognitive scores such as the Clinical Dementia Rating Sum of Boxes (CDR-SB) and and Mini Mental State Exam (MMSE) had especially high inter-site variance. These results are further discussed in Section \ref{disc:mci}. Feature importance rankings for all 6 DFNNs are presented in Fig. S6. We also examined the site-specific random intercepts of the ARMED-DFNN and found they correlated strongly with the percentage of pMCI subjects at each site (Pearson's $r = 0.860$, $p<10^{-5}$), indicating the random intercepts captured the variability in class balance across sites, a major source of confounding effect.

\begin{figure}[t]
    \centering
    \includegraphics[width=\columnwidth]{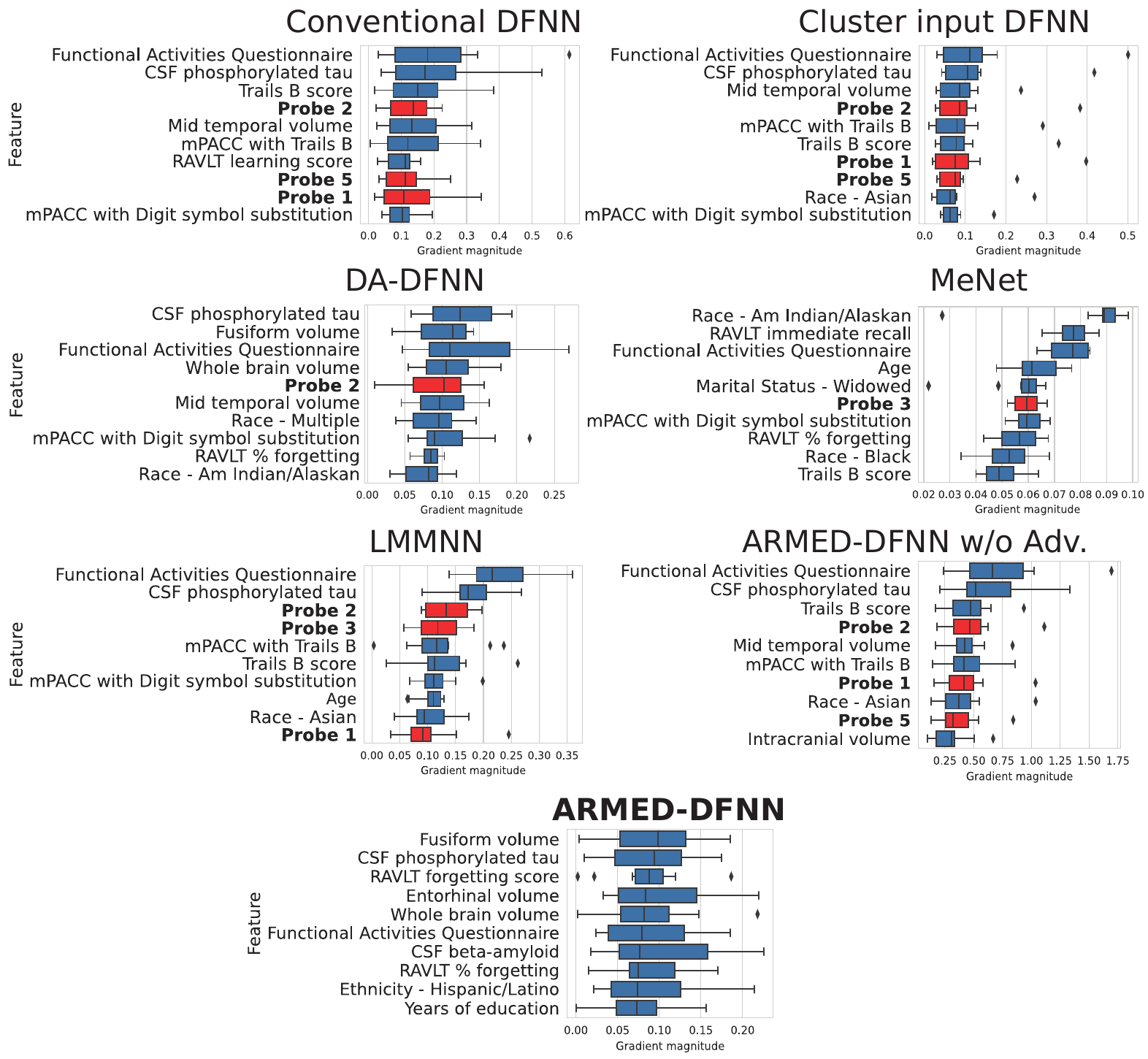}
    \caption{MCI conversion prediction with 5 added confounded probes (\textbf{bolded} label, red bar). For each DFNN, the top 10 features are shown, ranked by median feature importance (gradient magnitude) across 10 cross-validation folds.}
    \label{fig:mci_probes}
\end{figure}

When simulated confounded probe features were added, the ARMED-DFNN ranked these probes the lowest. The 10 highest ranked features for each model are shown in Fig. \ref{fig:mci_probes}. The conventional DFNN, cluster input DFNN, LMMNN, and ARMED-DFNN without domain adversarial regularization all included 3 of the 5 confounded probes within the top 10 features. The DA-DFNN and MeNet included 1 confounded probe and the full ARMED-DFNN included none in the top 10 features. Paired sign tests indicate that the ARMED-DFNN ranked the confounded probes significantly lower than any other model, e.g. $p = 0.031$ when compared to the second-best models, DA-DFNN and MeNet. 

\subsection{AD diagnosis}

\begin{table*}[!t]
    \renewcommand{\arraystretch}{1.25}
    \centering
    \begin{threeparttable}
    \caption{Alzheimer's Disease diagnosis from MRI. Metrics were computed through 10 Monte Carlo cross-validation replicates. Sensitivity and specificity were computed at the Youden point. The best results for each metric are \textbf{bolded}.}
    \label{table:ad}
    \begin{tabular}{@{\extracolsep{5pt}}l t r o r o r o r@{}}
    \hline
                      & \multicolumn{2}{c}{AUROC} & \multicolumn{2}{c}{Balanced accuracy (\%)} & \multicolumn{2}{c}{Sensitivity (\%)} & \multicolumn{2}{c}{Specificity (\%)} \\
    \cline{2-3} \cline{4-5} \cline{6-7} \cline{8-9}
    Model             & \dheader{Mean} & 95\% CI  & \dheader{Mean} & 95\% CI          & \dheader{Mean} & 95\% CI             & \dheader{Mean} & 95\% CI \\
    \hline
    \multicolumn{9}{c}{Seen sites} \\
    Conventional CNN            & 0.703         & 0.621 - 0.785   & 69.7         & 63.7 - 75.7  & 69.3         & 56.7 - 81.8  & 70.1          & 58.8 - 81.5 \\
    Cluster input CNN           & 0.654         & 0.585 - 0.722   & 72.6         & 68.2 - 76.9  & 76.0         & 67.4 - 84.7  & 69.1          & 56.5 - 81.8 \\
    DA-CNN                      & 0.823         & 0.730 - 0.917   & 79.9         & 74.2 - 85.6  & 77.2         & 61.4 - 93.0  & 82.6          & 76.8 - 88.5 \\
    MeNet                       & 0.923         & 0.894 - 0.952   & 89.6         & 86.7 - 92.5. & 87.7         & 82.0 - 93.4  & 91.5          & 88.9 - 94.2 \\
    LMMNN                       & \boldt{0.938} & 0.917 - 0.959   & \boldo{90.4} & 87.7 - 93.1  & 88.9         & 82.4 - 95.4  & \boldo{91.9}  & 88.4 - 95.3 \\
    ARMED-CNN                   & 0.900         & 0.861 - 0.939   & 88.7         & 83.8 - 91.6  & \boldo{91.8} & 87.0 - 96.7  & 83.6          & 75.2 - 92.0 \\
    \hspace{1em} w/o Adv.       & 0.816         & 0.729 - 0.903   & 79.7         & 73.3 - 86.2  & \boldo{91.6} & 86.1 - 97.2  & 67.8          & 52.6 - 83.1 \\
    \hspace{1em} randomized Z   & 0.585         & 0.506 - 0.664   & 63.3         & 58.3 - 68.2  & 65.5         & 48.3 - 82.7  & 61.0          & 47.2 - 74.8 \\
    \hline
    \multicolumn{9}{c}{Unseen sites} \\
    Conventional CNN            & 0.603         & 0.531 - 0.675   & 59.5         & 55.4 - 63.5  & 56.8         & 41.5 - 72.1  & 62.1          & 46.8 - 77.5 \\
    Cluster input CNN           & 0.587         & 0.520 - 0.653   & 58.3         & 54.4 - 62.2  & 57.6         & 42.3 - 72.9  & 59.0          & 42.9 - 75.2 \\
    DA-CNN                      & \boldt{0.652} & 0.614 - 0.690   & \boldo{62.5} & 59.7 - 65.3  & \boldo{71.8} & 66.7 - 77.0  & 53.1          & 48.9 - 57.4 \\
    MeNet                       & 0.517         & 0.463 - 0.571   & 53.9         & 51.6 - 56.3  & 64.9         & 45.7 - 84.1  & 43.0          & 23.6 - 62.3 \\
    LMMNN                       & 0.534         & 0.491 - 0.576   & 54.2         & 51.8 - 56.5  & 45.3         & 27.1 - 63.4  & \boldo{63.1}  & 45.7 - 80.6 \\
    ARMED-CNN                   & 0.645         & 0.606 - 0.684   & 61.2         & 58.6 - 63.9  & 65.9         & 60.3 - 71.6  & 56.6          & 50.8 - 62.3 \\
    \hspace{1em} w/o Adv.       & \boldt{0.655} & 0.608 - 0.701   & \boldo{62.2} & 58.9 - 65.6  & 68.6         & 61.6 - 75.6  & 55.9          & 46.6 - 65.1 \\
    \hspace{1em} randomized Z   & 0.551         & 0.526 - 0.576   & 54.6         & 53.0 - 56.2  & 47.5         & 33.9 - 61.1  & 61.8          & 47.4 - 76.1 \\
    \hline
    \end{tabular}
    \begin{tablenotes}
        \footnotesize
        \item CNN: convolutional neural network; DA: domain adversarial; Adv.: adversary; AUROC: area under receiver operating characteristic curve; CI: confidence interval. 
    \end{tablenotes}
    \end{threeparttable}
\end{table*}

\begin{figure}[h]
    \centering
    \includegraphics[width=0.9\columnwidth]{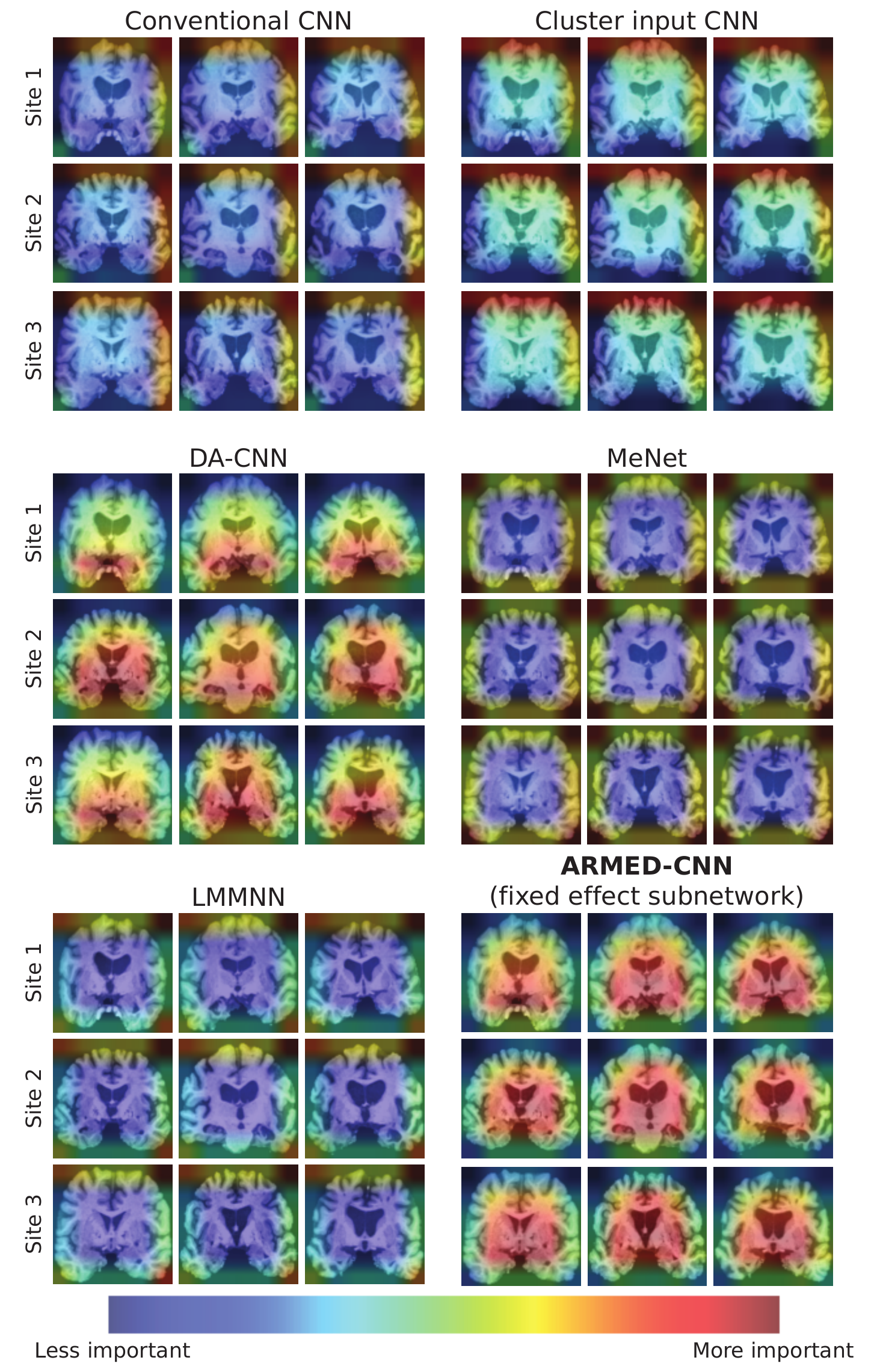}
    \caption{Grad-CAM visualizations indicating important image regions for classifying Alzheimer's Disease vs. cognitively normal individuals. Each row contains examples from one of three representative study sites.}
    \label{fig:gradcam}
\end{figure}

The cross-validated performance of each model in classifying brain MRIs as CN vs. AD is presented in Table \ref{table:ad}. On study sites seen during training, LMMNN showed the highest AUROC, followed by MeNet and the ARMED-CNN (Table \ref{table:ad}, top). Neither LMMNN nor MeNet significantly outperformed ARMED-CNN (paired T-test, $p = 0.064$ and $p = 0.210$, respectively). On unseen sites (Table \ref{table:ad}, bottom), the ARMED-CNN performed second best after the DA-CNN in AUROC, accuracy, and sensitivity. MeNet and LMMNN had the lowest AUROC on the unseen sites, indicating poor generalization. Without adversarial regularization, the performance of the ARMED-CNN increased slightly on unseen sites (mean AUROC 0.645 to 0.655) but decreased on seen sites (mean AUROC 0.900 to 0.816). Randomizing the site membership for seen sites drastically reduced all metrics, including mean AUROC from 0.900 to 0.585. On unseen sites, randomizing instead of inferring site membership reduced mean AUROC from 0.645 to 0.551.   

Gradient-weighted Class Activation Mapping (Grad-CAM) visualizations from each model revealed differences in the features learned (Fig. \ref{fig:gradcam}) \cite{Selvaraju.2020}. The conventional, cluster input, MeNet, and LMMNN CNNs attributed more weight to regions in the edges of each image, near the periphery of the brain. However, the DA-CNN emphasized medial brain areas, including the hippocampi and surrounding parahippocampal gyri. For the ARMED-CNN, we produced Grad-CAMs using the fixed effects subnetwork, which contains the learned cluster-invariant features. Like the DA-CNN, the ARMED-CNN also emphasized medial brain areas but gave additional weight to the superior regions including the lateral ventricles. Furthermore, we created separate Grad-CAMs to visualize the distinct site-specific features learned by the ARMED-CNN random effects subnetwork (Fig. S7), which involved the image periphery for some sites and more medial areas for others. 

\subsection{Cell image compression and classification}

\begin{table*}[t]
    \renewcommand{\arraystretch}{1.25}
    \centering
    \begin{threeparttable}
    \caption{Melanoma live cell image compression and classification, and batch effect contamination of latent representations. Confidence intervals were computed with DeLong's method. The best results for each metric are \textbf{bolded}.}
    \label{table:melanoma}
    \begin{tabular}{@{\extracolsep{5pt}}lfrrrtr@{}}
    \hline
    & \multicolumn{2}{c}{Seen batches} & \multicolumn{2}{c}{Unseen batches} & \multicolumn{2}{c}{Latent batch clustering} \\
    \cline{2-3} \cline{4-5} \cline{6-7}
    Model       & \dheader{MSE}  & \multicolumn{1}{c}{AUROC ({\scriptsize 95\% CI})} & \dheader{MSE} & \multicolumn{1}{c}{AUROC ({\scriptsize 95\% CI})} & \dheader{DB}  & \dheader{CH} \\ 
    \hline
    Conventional AEC            & 0.0019         & 0.817 ({\scriptsize 0.812 - 0.822})          & 0.0024     & 0.773 ({\scriptsize 0.764 - 0.781})          & 8.885          & 545.9 \\
    DA-AEC                      & 0.0018         & 0.777 ({\scriptsize 0.771 - 0.783})          & 0.0024     & 0.759 ({\scriptsize 0.750 - 0.768})          & \boldt{43.009} & \textbf{20.4} \\
    ARMED-AEC                   & \boldf{0.0012} & 0.869 ({\scriptsize 0.865 - 0.874})          & 0.0024     & \textbf{0.789} ({\scriptsize 0.781 - 0.798}) & \boldt{43.009} & \textbf{20.4} \\ 
    \hspace{1em} w/o Adv.       & \boldf{0.0012} & \textbf{0.876} ({\scriptsize 0.872 - 0.881}) & 0.0024     & \textbf{0.791} ({\scriptsize 0.782 - 0.799}) & 8.885          & 545.9 \\
    \hspace{1em} randomized Z   & 0.0018         & 0.732 ({\scriptsize 0.726 - 0.738})          & 0.0024     & 0.712 ({\scriptsize 0.702 - 0.721}) & & \\
    \hline
    \end{tabular}
    \begin{tablenotes}
        \footnotesize
        \item AEC: autoencoder-classifier; DA: domain adversarial; Adv.: adversary; MSE: mean squared error between original and reconstructed images; AUROC: area under receiver operating characteristic curve for phenotype classification; CI: confidence interval; DB: Davies-Bouldin score, lower values indicate stronger clustering; CH: Calinski-Harabasz score, higher values indicate stronger clustering
    \end{tablenotes}
    \end{threeparttable}
\end{table*}

The performance of each AEC model in compressing and classifying melanoma live-cell images is presented in Table \ref{table:melanoma}. For computational efficiency, the pre-trained and frozen DA-AEC was reused as the fixed effects subnetwork of the ARMED-AEC. For the ablation test without adversarial regularization (``w/o Adv."), the pre-trained conventional AEC was reused as the fixed effects subnetwork. Confidence intervals were computed using DeLong's method \cite{DeLong.1988}. On \textit{seen} batches, (Table \ref{table:melanoma}, first column group) the ARMED-AEC had the highest performance in classifying metastatic efficiency (AUROC 0.869), statistically significantly outperforming the second-best model, the conventional AEC ($p < 0.001$), and it had the lowest reconstruction error (MSE 0.0012). On \textit{unseen} batches (Table \ref{table:melanoma}, second column group), the ARMED-AEC again showed the best classification performance (AUROC 0.789). This classification performance was statistically significantly higher than the second-best model, the conventional AEC ($p < 0.001$). All models had similar reconstruction error (MSE 0.0024) on unseen batches. Examining each AEC's latent representations, the DA-AEC and ARMED-AEC (using the DA-AEC as its fixed effects subnetwork) exhibited much less batch effect contamination. Compared to the conventional AEC, the DB score improved from 8.885 to 43.009 (484\% relative increase) and the CH score improved from 545.9 to 20.4 (96\% relative decrease). 

In the ablation tests, removing the domain adversarial regularization of the fixed effects subnetwork in the ARMED-AEC (using the conventional AEC as the fixed effects subnetwork) slightly increased classification AUROC on \textit{seen} batches (0.876 vs. 0.869) and on unseen batches (0.791 vs. 0.789). However, this came at the expense of greater batch contamination of the latent representations (DB score 8.885 and CH score 545.9). When cluster assignments were randomized instead of using the true cluster assignments on seen batches, reconstruction MSE worsened from 0.0012 to 0.0018 and classification AUROC decreased from 0.869 to 0.732. On \textit{unseen} batches, randomized instead of Z-predictor-inferred cluster assignments reduced classification AUROC from 0.789 to 0.712. 

\begin{figure}[t]
    \centering
    \includegraphics[width=0.625\columnwidth]{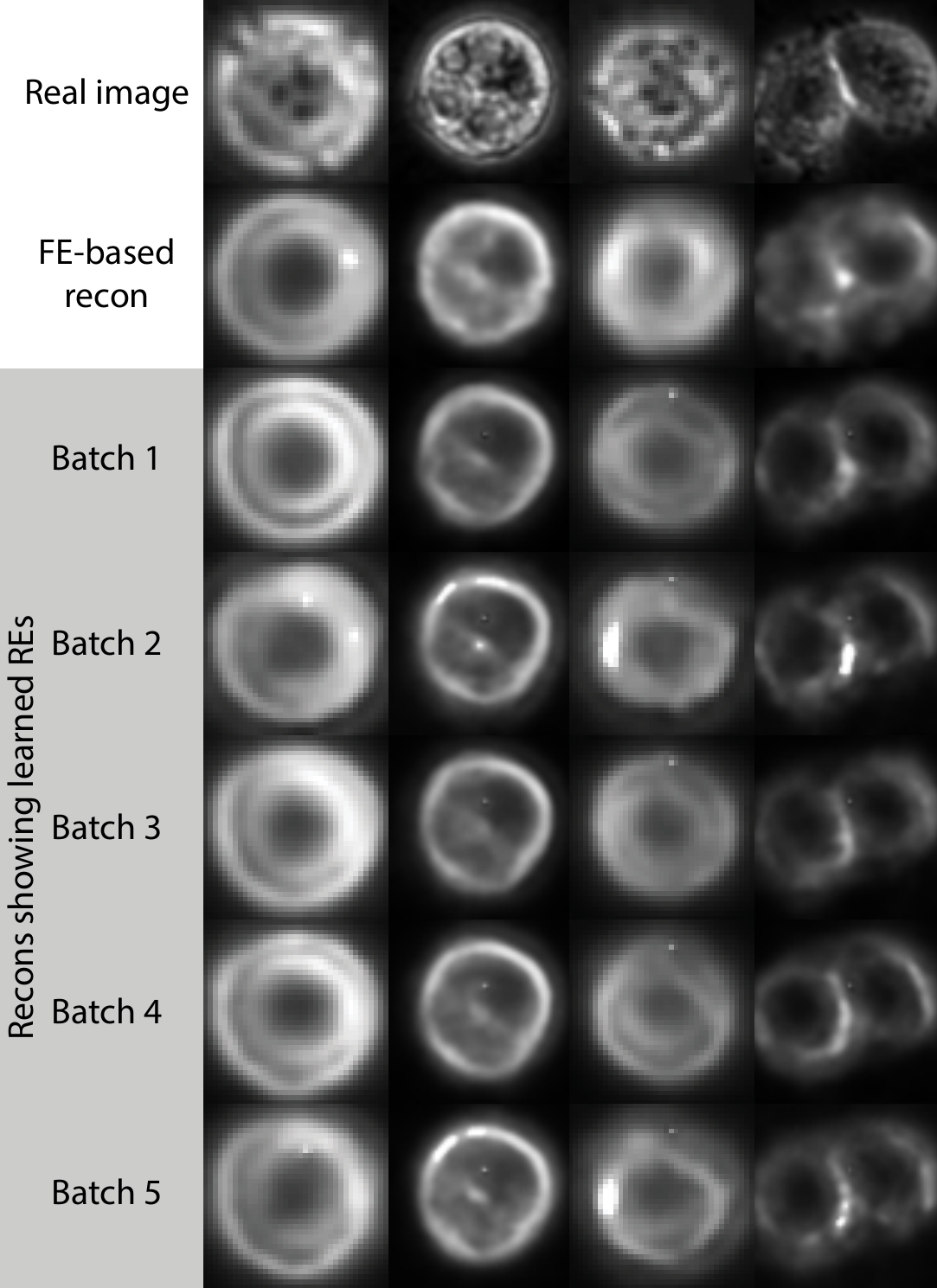}
    \caption{Reconstructed melanoma cell images from the ARMED-AEC. The first row contains the real image, the second row contains the fixed effects-based reconstructions, and remaining rows show random effects-based reconstructions using different learned random effects.}
    \label{fig:melanoma_recons}
\end{figure}

To visualize the random effects learned by the ARMED-AEC, we generated image reconstructions from the random effects subnetwork with various learned batch-specific effects applied (Fig. \ref{fig:melanoma_recons}). These simulate the appearance of an image if it had been acquired within different batches. We compared these with the image reconstructions from the fixed effects subnetwork, where batch effects have been removed. Some batches showed stronger specular highlights (e.g. batches 2 and 5), while others had greater contrast in the cell periphery (e.g. batches 1 and 3).

\section{Discussion}

\subsection{General observations}
Our experiments across four applications illustrate the three critical contributions of ARMED. First, we demonstrated that the fixed effects subnetwork of ARMED models assigns feature importance more appropriately than the compared models. In the spiral simulations, the ARMED-DFNN most strongly separated the true and confounded features by feature importance. The conventional and MLDG models erroneously placed greater importance on the confounded probes than the true features, while the cluster input, DA-DFNN, MeNet, and LMMNN models downweighted the confounded probes to a lesser degree than the ARMED-DFNN. In MCI conversion prediction with simulated confounded probes, the ARMED-DFNN ranked the probes statistically significantly lower than any other model, including the DA-DFNN and MeNet. In contrast, the conventional, cluster input, and LMMNN models were most sensitive to the probes. In AD diagnosis, Grad-CAM visualizations showed that the ARMED-CNN highlighted more biologically plausible brain regions than the conventional, cluster input, MeNet, and LMMNN CNN's, which is further discussed in Section \ref{disc:ad}. 

Second, we demonstrated the ability of ARMED to visualize random effects learned by the random effects subnetwork. In MCI conversion prediction, we quantified the learned inter-site variance of the random slopes for each feature. This allowed us to identify which features are most contaminated by site effects, and we discuss these below (Section \ref{disc:mci}). In AD diagnosis, we visualized site-specific differences in Grad-CAMs. Finally, in the cell imaging application, we generated image reconstructions showing the impact of learned batch effects.

Third, ARMED typically outperforms the compared non-mixed effects methods and outperforms or matches the other mixed effects methods. In the spiral simulations, the ARMED-DFNN had either the best or second-best accuracy, while being more discriminative between true and confounded features and learning the most cluster-appropriate decision boundaries. In the MCI conversion application, the ARMED-DFNN outperformed all other methods on both data from seen and unseen sites.  In AD diagnosis, the ARMED-CNN performed similarly to MeNet and LMMNN methods on seen sites and competed favorably with the DA-CNN on unseen sites. Meanwhile, MeNet and LMMNN generalized poorly to unseen sites. In the cell imaging application, the ARMED-AEC had the best reconstruction error on data from seen batches, the best metastatic efficiency classification on both seen and unseen batches, and substantially reduced batch effects in its latent representations compared to the conventional AEC. In ablation tests, we found that ARMED models without DA often performed similarly to or non-significantly better than the full model with DA, but their fixed effects subnetworks were more sensitive to confounded probe features. Therefore, we recommend always using the full ARMED model with DA, as any small performance increase comes at the cost of confound susceptibility. We also found that randomizing cluster assignment reduced performance on seen clusters, confirming that the ARMED models had learned cluster-specific information in the random effects subnetworks. Similarly, performance on unseen clusters decreased when using randomized cluster assignment instead of using the Z-predictor to infer cluster membership. This indicates that the Z-predictor is needed to fully exploit the learned random effects when predicting on data from unseen clusters. 

Though we have focused on biomedical data in this work, we anticipate that our approach will be of use to any case where data is non-\textit{iid} and subject to random effects. Given its flexible and modular nature, the ARMED framework should apply readily to other architecture types besides the three demonstrated here. 

\subsection{Comparison to prior work}
A common approach to handling clustered data is to include the cluster membership, which is an unordered categorical variable, as additional one-hot encoded covariates in $X$ \cite{Hancock.2020}. This approach is unable to disentangle the cluster-specific random effects and cluster-independent fixed effects, and we found it was more sensitive to simulated confounded probes than ARMED models. We also found inferior performance vs. ARMED, likely due to the high cardinality of the added features which can lead to overfitting \cite{Rao.2017, Simchoni.2021}. For example, the MCI conversion application had 20 sites and 37 input features, meaning that to add cluster membership to $X$ would increase the width of $X$ by 35\%. ARMED is better suited to handling this high-cardinality information by modeling clustering as a random effect, which imposes a normal distribution prior. 

A more recent approach to handling differences across clusters is domain adversarial learning. We showed that DA does improve generalization to data from unseen clusters. However, ARMED improves upon DA, adding a random effects subnetwork to capture the cluster-specific information that DA discards, which results in better performance on clusters \textit{seen} during training. Using the Z-predictor, this cluster-specific information can also be used when predicting on data from unseen clusters, allowing ARMED to outperform DA on \textit{unseen} clusters as well. 

This work remedies key weaknesses in previous approaches to incorporate mixed effects into deep learning. We described specific random effects architectures for random intercepts, linear slopes, and/or nonlinear slopes. This allows greater flexibility than DeepGLMM and LMMNN, which only learn random intercepts, and MeNet, which only learns nonlinear random slopes \cite{Xiong.2019, Tran.2020, Simchoni.2021}. Another key improvement was adversarial regularization of the fixed effects subnetwork to learn generalizable, cluster-agnostic information. In our experiments with simulated confounders, this allowed ARMED models to appropriately upweight nonconfounded features and downweight confounded features, while MeNet and LMMNN, lacking adversarial regularization, were susceptible to the spurious confounded features. Next, we demonstrated interpretation and visualization of the learned random effects, which was not explored in these previous works. Finally, we evaluated ARMED models on data from clusters unseen during training and provided a method to infer cluster membership and apply learned random effects on this data. The previous works lacked such a method, meaning that the learned random effects cannot be utilized on new data. This is a major limitation for practical applications, where a deployed model may need to be applied to data from a new cluster, such as a new clinical site or patient. 

\subsection{Application-specific discussions}

\subsubsection{MCI conversion prediction} \label{disc:mci}
The ARMED-DFNN quantifies the inter-site variance of the learned random slope for each feature (Fig. \ref{fig:mci_me_features}). We found that demographic features such as race and ethnicity had the lowest inter-site variance, which in unsurprising as the association between these features and MCI conversion should not be sensitive to measurement differences across sites. Certain cognitive measurements, however, had distinctly high inter-site variance. The CDR-SB score had the highest variance, which concurs with a previous report that CDR-SB has suboptimal inter-rater reliability in early dementia patients, such as those with MCI \cite{Rockwood.2000}. MMSE had the second highest variance in our ARMED-DFNN, again agreeing with previous findings of low inter-rater reliability \cite{Bowie.1999}.

Though we intentionally held out a large portion of the ADNI dataset to evaluate our models on unseen sites, our ARMED-DFNN performed similarly to or better than several published results on predicting 24-month MCI conversion in ADNI using deep learning. Lee et al. achieved 80\% accuracy compared to the 81.9\% of our ARMED-DFNN \cite{Lee.2019}. Shi et al. and Lian et al. achieved AUROC of 0.816 and 0.793, respectively, compared to our 0.926 \cite{Shi.2021, Lian.2020}. Note that neither of these studies held out entire study sites for evaluation, and our AUROC on \textit{unseen} sites (0.837) still exceeded their results on \textit{seen} sites. 

\subsubsection{AD diagnosis} \label{disc:ad}
The Grad-CAMs of the DA-CNN and ARMED-CNN appropriately emphasized the importance of medial brain regions including the hippocampus and surrounding medial temporal lobe, which are involved in AD-related brain atrophy (Fig. \ref{fig:gradcam}) \cite{Schott.2005, Ferrarini.2006, Dubois.2014}. The ARMED-CNN Grad-CAMs also indicated the importance of the lateral ventricles, where enlargement has been connected to AD \cite{Ferrarini.2006, Apostolova.2012}. The incorporation of these additional structures likely contributed to the better performance of the ARMED-CNN (AUROC 0.900) vs. the DA-CNN (AUROC 0.823). Meanwhile, the conventional, cluster input, MeNet, and LMMNN models relied highly on likely spurious features in the image periphery. Such features appear to be related to site effects on imaging, since the random effects of the ARMED-CNN affect similar peripheral areas (Fig. S7).

The performance of our ARMED-CNN compares favorably to previous models using 2D MRI to diagnose AD in the ADNI dataset. We achieved 88.7\% accuracy, while Kang et al. report 90.4\% and Ebrahimi et al. report 87.5\% \cite{Kang.2021, Ebrahimi.2021}. However, we trained on a fraction of the total ADNI data that these reports used, holding out the rest for evaluation of models on unseen sites. Consequently, our work focuses on comparisons across architectures, not with previous studies. 

\subsubsection{Cell image compression and classification} \label{disc:melanoma}
We compare our ARMED-AEC results to the previous analysis published by Zaritsky et al. \cite{Zaritsky.2021}. While they discussed the batch effect present in the latent representations produced by their autoencoder, their methods did not explicitly suppress this batch effect. In contrast, our proposed ARMED-AEC reduced the batch effect in the latent representations by 484\% based on the DB score, compared to an unmodified AEC. We also improved classification AUROC to 0.876 compared to their reported 0.723, though this may be partially due to the direct incorporation of the phenotype classifier into the autoencoder, while Zaritsky et al. trained their classifier separately from their autoencoder. 

\subsection{Limitations}
Mixed effects models generally require the presence of several clusters to accurately estimate the random effect distributions; with \textless 4 clusters, LME models provide less of an advantage over generalized linear regression \cite{Gelman.2007, Harrison.2018}. Consequently, we suggest some caution when using our method for data with fewer than 4 clusters. Additionally, our method applies to datasets with a single level of random effects, but there are often cases with multiple levels of random effects, such as when multiple observations are collected per subject who are then clustered by study site. We plan to extend our methodology to such multi-level cases in future work. Finally, a practical limitation of ARMED is the additional complexity, which may increase training time by approximately 1.5-2x. However, we note that other methods have even greater computation cost, such as meta-learning domain generalization (MLDG) which uses second-order optimization and MeNet and LMMNN which involve expensive matrix inversions \cite{Li.2018, Xiong.2019, Simchoni.2021}.



\section{Conclusion}

Our proposed approach uses powerful mixed effects techniques from traditional statistics to improve the interpretability, reliability, and performance of deep learning models on non-\textit{iid} data. ARMED models separately learn random effects and fixed effects in distinct subnetworks, with the fixed effects subnetwork more appropriately assigning feature importance with resilience to confounding effects, helping to avoid Type I and Type II errors. In biomedical applications, this allows better hypothesis formation and prevents waste of resources in following up biased or confounded results. Meanwhile, the random effects subnetwork allows users to understand the cluster effects in their data, which can inform future research. For example, clinical study organizers could prioritize measurements with less inter-site variance in future studies. Besides these benefits, ARMED also increases predictive performance on clustered data, including better generalization to clusters unseen during training. Given these advantages demonstrated across multiple model architectures and applications, we broadly recommend the ARMED framework to deep learning practitioners dealing with non-\textit{iid} data. We make our code available at \texttt{\url{tinyurl.com/ARMEDCode}}.

\ifCLASSOPTIONcompsoc
  \section*{Acknowledgments}
\else
  \section*{Acknowledgment}
\fi
\footnotesize
Data collection and sharing for this project was funded by the Alzheimer's Disease Neuroimaging Initiative (ADNI) (National Institutes of Health Grant U01 AG024904) and DOD ADNI (Department of Defense award number W81XWH-12-2-0012). ADNI is funded by the National Institute on Aging, the National Institute of Biomedical Imaging and Bioengineering, and through generous contributions from the following: AbbVie, Alzheimer’s Association; Alzheimer’s Drug Discovery Foundation; Araclon Biotech; BioClinica, Inc.; Biogen; Bristol-Myers Squibb Company; CereSpir, Inc.; Cogstate; Eisai Inc.; Elan Pharmaceuticals, Inc.; Eli Lilly and Company; EuroImmun; F. Hoffmann-La Roche Ltd and its affiliated company Genentech, Inc.; Fujirebio; GE Healthcare; IXICO Ltd.; Janssen Alzheimer Immunotherapy Research \and Development, LLC.; Johnson \and Johnson Pharmaceutical Research \and Development LLC.; Lumosity; Lundbeck; Merck \and Co., Inc.; Meso Scale Diagnostics, LLC.; NeuroRx Research; Neurotrack Technologies; Novartis Pharmaceuticals Corporation; Pfizer Inc.; Piramal Imaging; Servier; Takeda Pharmaceutical Company; and Transition Therapeutics. The Canadian Institutes of Health Research is providing funds to support ADNI clinical sites in Canada. Private sector contributions are facilitated by the Foundation for the National Institutes of Health (www.fnih.org). The grantee organization is the Northern California Institute for Research and Education, and the study is coordinated by the Alzheimer’s Therapeutic Research Institute at the University of Southern California. ADNI data are disseminated by the Laboratory for Neuro Imaging at the University of Southern California.

We thank the lab of Dr. Gaudenz Danuser at the University of Texas Southwestern Medical Center for providing the melanoma live-cell microscopy dataset, Dr. Andrew Jamieson for technical advice and manuscript feedback, and Dr. Jian Zhou for manuscript feedback.

\ifCLASSOPTIONcaptionsoff
  \newpage
\fi



\bibliographystyle{IEEEtran}
\bibliography{references}

%
\vspace{-3em}

\begin{IEEEbiography}[{\includegraphics[width=1in,height=1.25in,clip,keepaspectratio]{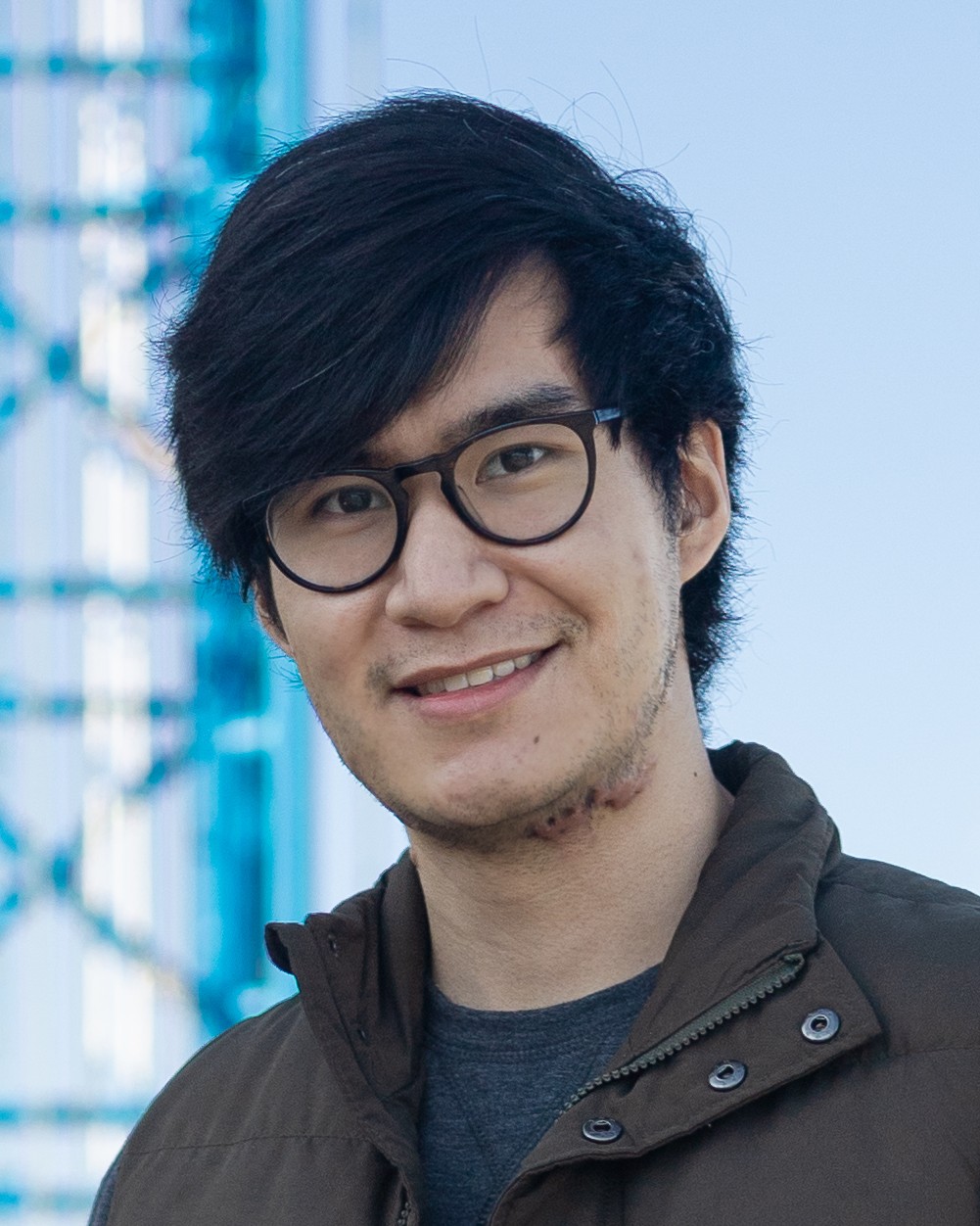}}]{Kevin P. Nguyen}
received his BS in biomedical engineering from Yale University. He is currently enrolled in the Medical Scientist Training Program at UT Southwestern, working toward both a medical doctorate (MD) and a PhD in biomedical engineering. His interests include the development of deep learning techniques to improve interpretability and performance on medical applications, especially in neuroradiology. 
\end{IEEEbiography}

\begin{IEEEbiography}[{\includegraphics[width=1in,height=1.25in,clip,keepaspectratio]{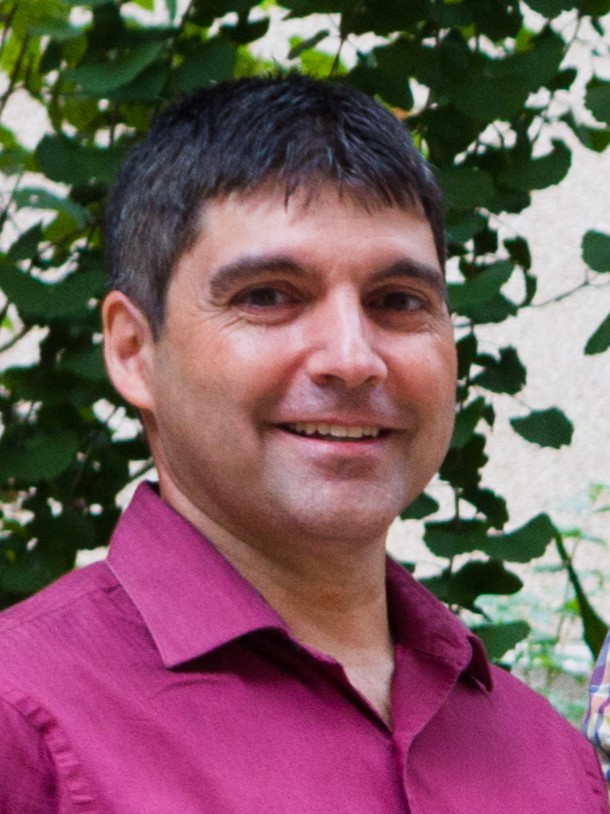}}]{Albert A. Montillo}
received his BS and MS in computer science from Rensselaer University and PhD in medical imaging from the University of Pennsylvania. He is an Assistant Professor in the Lyda Hill Dept. of Bioinformatics at UT Southwestern and the head of the Deep Learning for Precision Health Lab. His work emphasizes algorithmic developments in machine learning to integrate brain imaging with multimodal patient data, improving diagnoses, prognoses, and treatment planning for neurological disorders. 
\end{IEEEbiography}





\end{document}


\maketitle
\tableofcontents

\section{Supplemental tables}

\begin{table}[ht]
\caption{Characteristics and MRI acquisition parameters for ADNI2 and ADNI3 study sites included in Alzheimer's Disease diagnosis training dataset.}
\label{table:adni_scanners}
\centering
\resizebox{\textwidth}{!}{
    \begin{tabular}{@{}rrrrllrrrr@{}}
    \toprule
    Site ID & Subjects & Images & \% AD & Manufacturer         & Model   & Flip angle   & TE (ms) & TI (ms) & Resolution (mm) \\ \midrule
    52      & 8        & 14     & 85.7  & General Electric     & Discovery MR750 & 11\degree  & 3      & 400         & $1.2 \times 1.1 \times 1.1$ \\
    5       & 7        & 16     & 81.3  & General Electric     & Discovery MR750 & 11\degree  & 3      & 400         & $1.2 \times 1.0 \times 1.0$ \\
    126     & 12       & 26     & 76.9  & General Electric     & Discovery MR750 & 11\degree  & 3      & 400         & $1.2 \times 1.0 \times 1.0$ \\
    57      & 6        & 9      & 55.6  & General Electric     & Discovery MR750 & 11\degree  & 3      & 400         & $1.2 \times 1.0 \times 1.0$ \\
    16      & 21       & 56     & 50.0  & General Electric     & Signa HDxt      & 11\degree  & 3      & 400         & $1.2 \times 1.0 \times 1.0$ \\
    2       & 24       & 80     & 11.3  & Philips              & Intera          & 9\degree   & 3      & 900         & $1.2 \times 1.1 \times 1.1$ \\
    100     & 15       & 28     & 7.1   & Philips              & Achieva         & 9\degree   & 3      & 900         & $1.0 \times 1.0 \times 1.0$ \\
    73      & 16       & 56     & 8.9   & Siemens              & TrioTim         & 9\degree   & 3      & 900         & $1.2 \times 1.1 \times 1.1$ \\
    41      & 27       & 79     & 3.8   & Siemens              & TrioTim         & 9\degree   & 3      & 900         & $1.2 \times 1.1 \times 1.1$ \\
    22      & 9        & 30     & 3.3   & Siemens              & TrioTim         & 9\degree   & 3      & 900         & $1.2 \times 1.1 \times 1.1$ \\
    941     & 29       & 66     & 1.5   & Siemens              & Prisma Fit      & 9\degree   & 3      & 900         & $1.0 \times 1.1 \times 1.1$ \\
    20      & 10       & 22     & 0.0   & Siemens              & Prisma Fit      & 9\degree   & 3      & 900         & $1.0 \times 1.0 \times 1.0$ \\ \bottomrule
    \end{tabular}
}
\begin{tablenotes}
\footnotesize
\item AD: Alzheimer's Disease; TE: echo time; TI: inversion time
\end{tablenotes}
\end{table}

\newpage
\section{Supplemental figures}

\begin{figure}[h]
    \centering
    \includegraphics[scale=0.6]{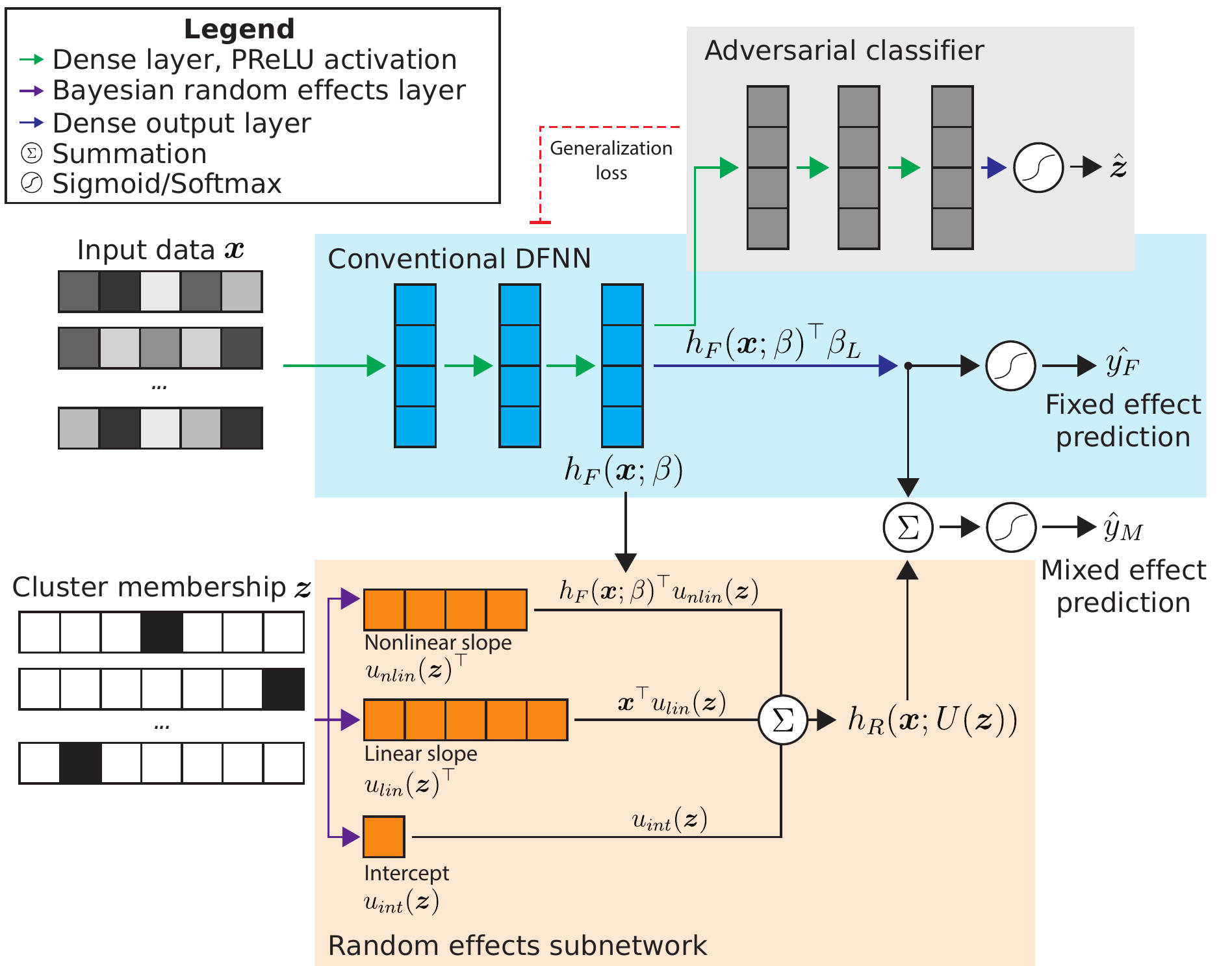}
    \caption{ARMED dense feedforward neural network (ARMED-DFNN). A conventional DFNN (blue area) applies a nonlinear transformation to the input data sample $\boldsymbol{x}$, yielding $h_F(\boldsymbol{x}, \beta)$ as the output from the penultimate dense layer. The final dense layer with weights $\beta_L$ then produces a fixed effects-only prediction $\hat{y}_F$. The fixed effects subnetwork consists of this conventional DFNN and an adversarial classifier (gray area) which, through the generalization loss, penalizes the DFNN for learning features that are predictive of the sample's cluster $\hat{\boldsymbol{z}}$. The random effects subnetwork (orange area) learns normally-distributed weights $U(\boldsymbol{z}) \sim N(0, \Sigma)$ dependent on cluster $\boldsymbol{z}$. These include nonlinear slopes of $h_F(\boldsymbol{x}, \beta)$, linear slopes of $\boldsymbol{x}$, and/or intercepts. The combination (e.g. sum) of the fixed and random effects subnetwork outputs produces the mixed effects-based prediction $\hat{y}_M$.}
    \label{fig:dfnn_architecture}
\end{figure}

\begin{figure}[ht!]
    \centering
    \includegraphics[scale=0.8]{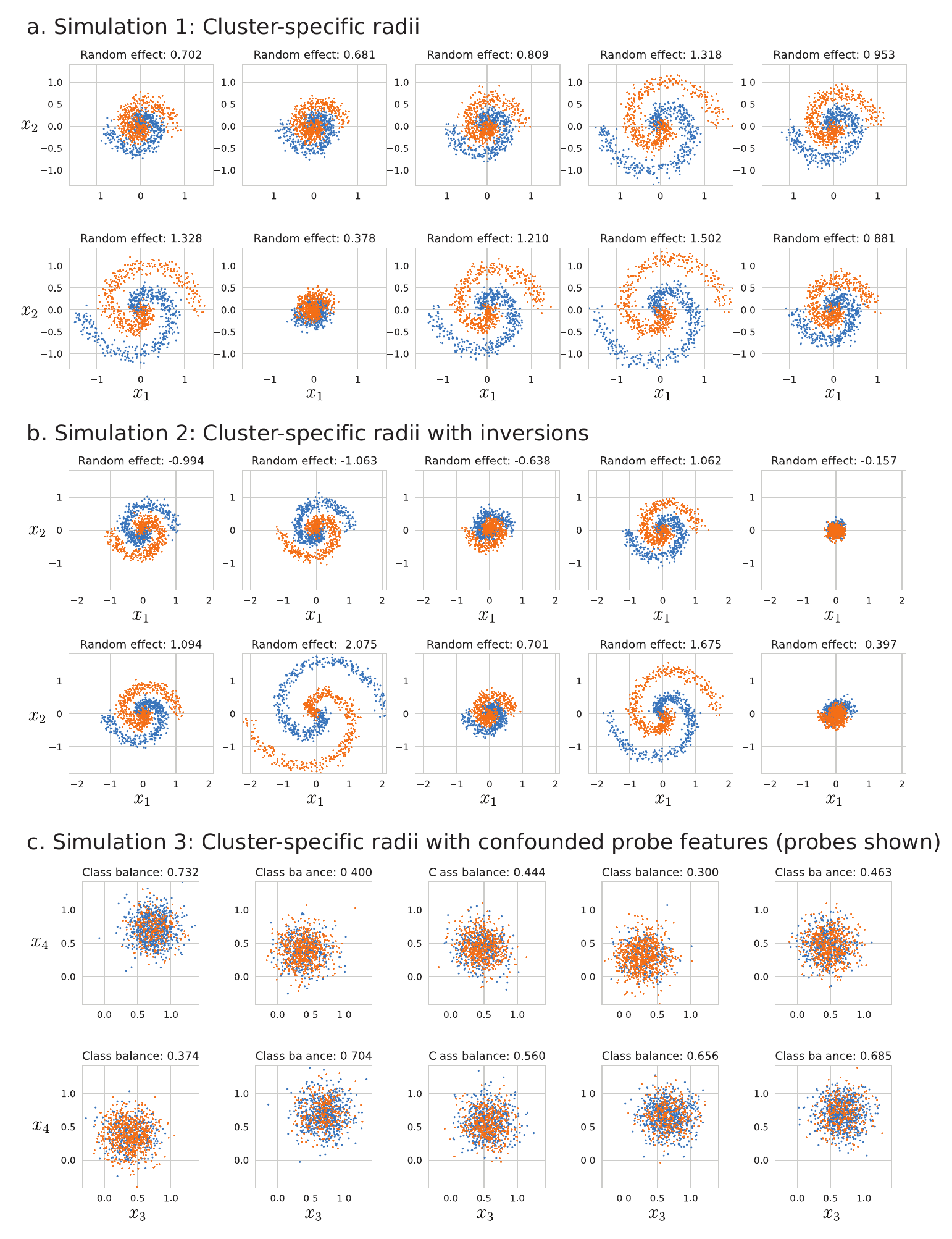}
    \caption{Spiral simulation datasets. Each simulated point has two features, $x_1$ and $x_2$, and the prediction target is whether it belongs to the blue or orange spiral. a) In simulation 1, 10 clusters were generated with cluster-specific spiral radii sampled from a normal distribution with $\mu = 1$ and $\sigma = 0.3$. b) In simulation 2, the cluster-specific radii were sampled from a normal distribution with $\mu = 0$ and $\sigma = 1$ such that spiral labels were inverted when the radius $< 0$. c) Simulation 3 was generated using the same parameters as simulation 1. However, two confounded probe features $x_3$ and $x_4$ were added, shown here, which were correlated with the ratio of blue-to-orange points but unassociated with the underlying spiral function.}
    \label{fig:spiral_data}
\end{figure}

\begin{figure}[t]
    \centering
    \includegraphics[scale=0.65]{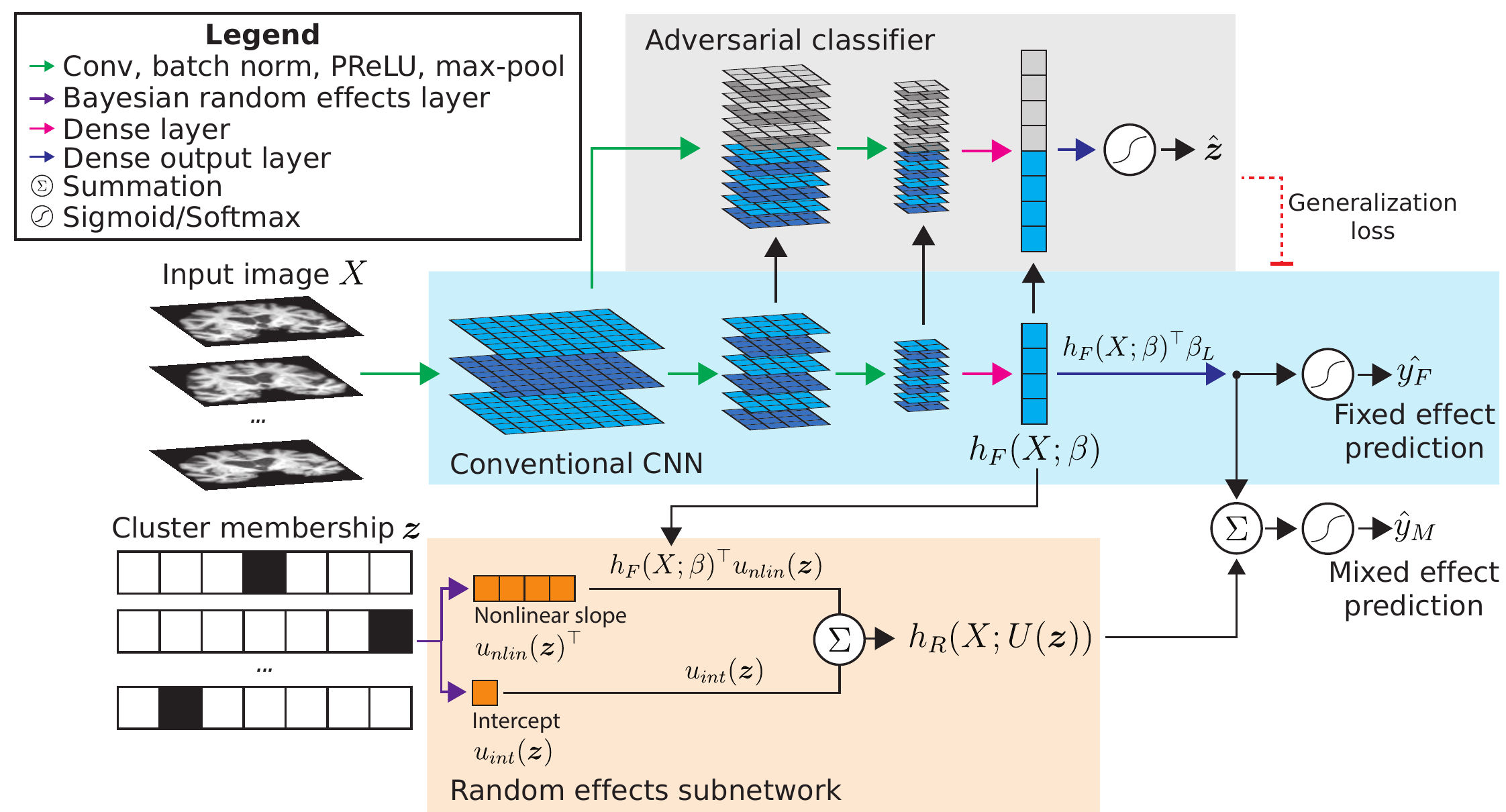}
    \caption{ARMED convolutional neural network (ARMED-CNN). For the conventional CNN (blue area), the penultimate dense layer produces a latent representation $h_F(X, \beta)$ of the data $X$. A final dense layer with weights $\beta_L$ produces the classification prediction $\hat{y}_F$. To create the fixed effects subnetwork, an adversarial classifier (gray area) is added that learns to predict the sample's cluster $\hat{z}$ from the outputs of the layers of the CNN. The generalization loss penalizes the CNN for learning features that allow the adversarial classifier to predict cluster $\hat{z}$. The random effects subnetwork (orange area) learns normally-distributed weights $U(Z) \sim N(0, \Sigma)$ dependent on cluster $\hat{z}$, including scalars (nonlinear slopes) applied to $h_F(X, \beta)$ and a bias. These are combined with the fixed effects subnetwork output to yield the mixed effects-based prediction $\hat{y}_M$. Note: for illustration, not all CNN layers are shown.}
    \label{fig:cnn_architecture}
\end{figure}

\begin{figure}[p]
    \centering
    \includegraphics[scale=0.57]{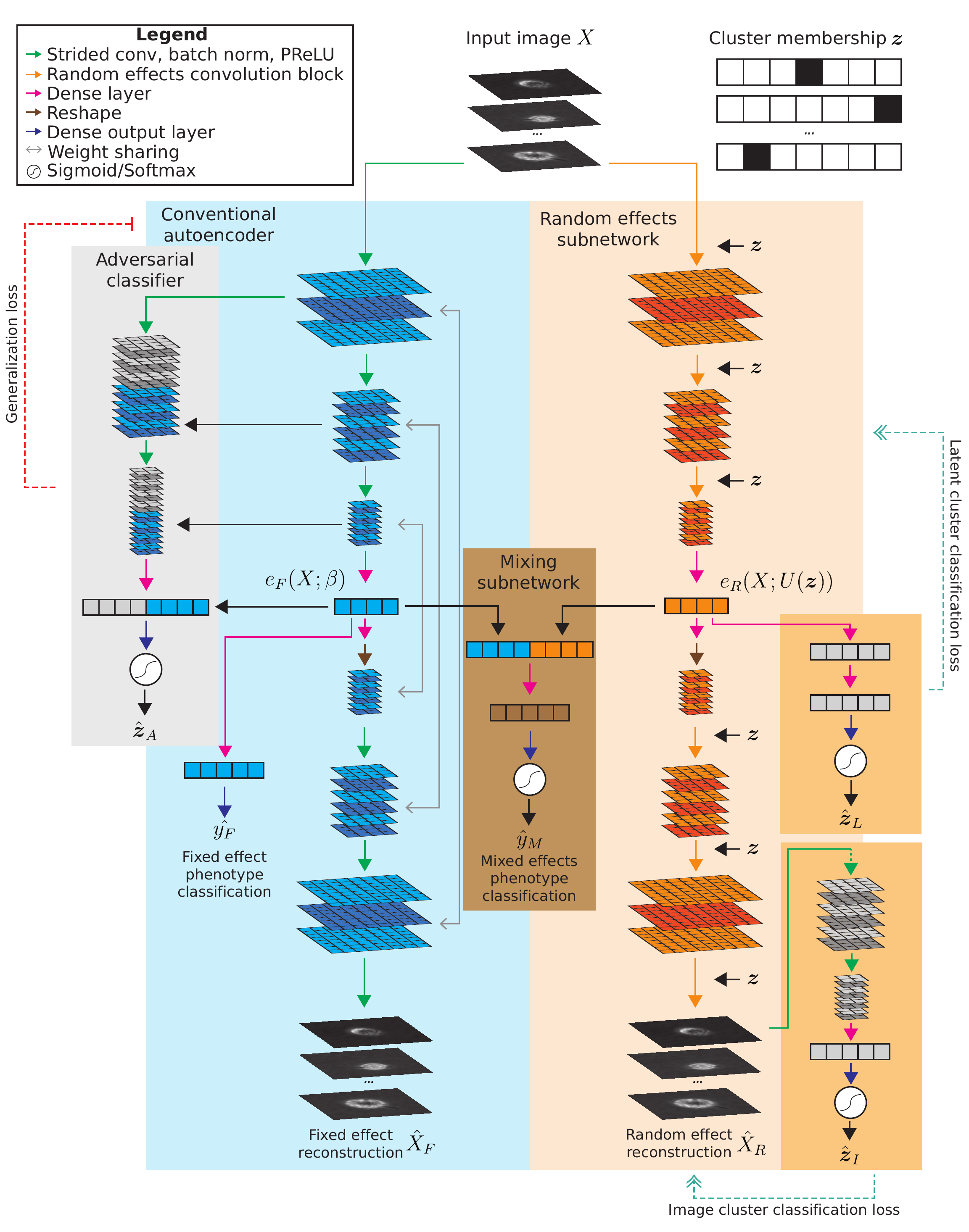}
    \caption{ARMED autoencoder-classifier (ARMED-AEC). A conventional autoencoder-classifier (blue area) contains an encoder that compresses the input image $X$ into a latent representation $e_F(X; \beta)$, an auxiliary classifier that predicts the phenotype label $\hat{y}_F$, and a decoder that reconstructs the image $\hat{X}_F$. The fixed effects subnetwork combines this conventional model with an adversarial classifier (gray area), which penalizes the encoder through the generalization loss for learning features predictive of cluster membership $\hat{\boldsymbol{z}}_A$. The decoder shares weights with the encoder and so receives the same adversarial guidance. Meanwhile, the random effects subnetwork (orange areas) is a parallel autoencoder using random effects convolution blocks with cluster-dependent weights (Fig. S5). Its encoder compresses the image into a latent representation $e_R(X; U(\boldsymbol{z}))$ containing information predictive of cluster $\hat{\boldsymbol{z}}_L$. Its decoder then reconstructs an image $\hat{X}_R$ that is also contains information predictive of cluster $\hat{\boldsymbol{z}}_I$. A learned combination of the two latent representations yields the mixed effects-based phenotype prediction $\hat{y}_M$. Note: for illustration, not all autoencoder layers are shown.}
    \label{fig:aec_architecture}
\end{figure}

\begin{figure}[ht!]
    \centering
    \includegraphics{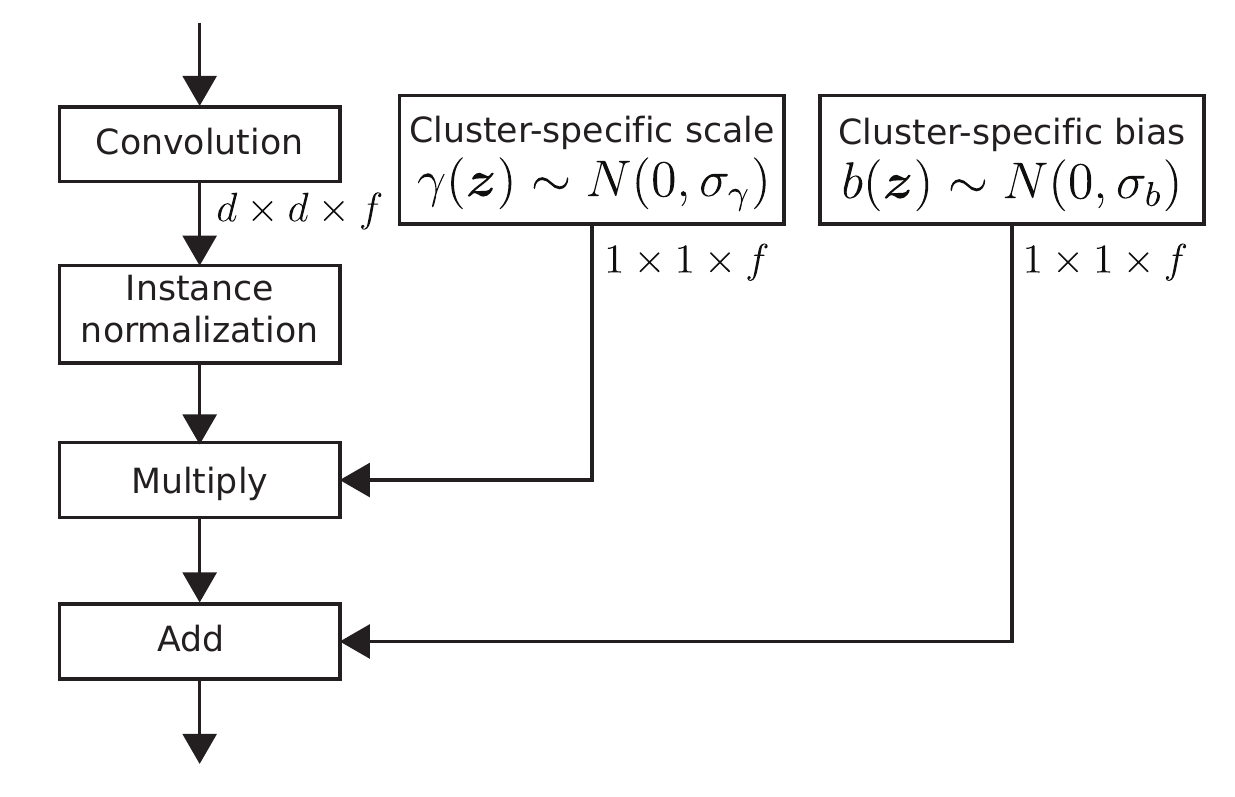}
    \caption{Random effects convolution (RECON) block used in the proposed mixed effects autoencoder-classifier (ARMED-AEC). A convolutional layer with $f$ filters yields a $d \times d \times f$ output tensor. Instance normalization is applied, centering and rescaling values to zero mean and unit variance. A cluster-specific scale $\gamma(\boldsymbol{z})$ and bias $b(\boldsymbol{z})$ are then applied to each of the $f$ feature maps. These cluster-specific scales and biases are regularized to follow normal distributions with learned variances $\sigma_\gamma$ and $\sigma_b$, respectively.}
    \label{fig:recon_block}
\end{figure}

\begin{figure}[p]
    \centering
    \includegraphics[width=\textwidth]{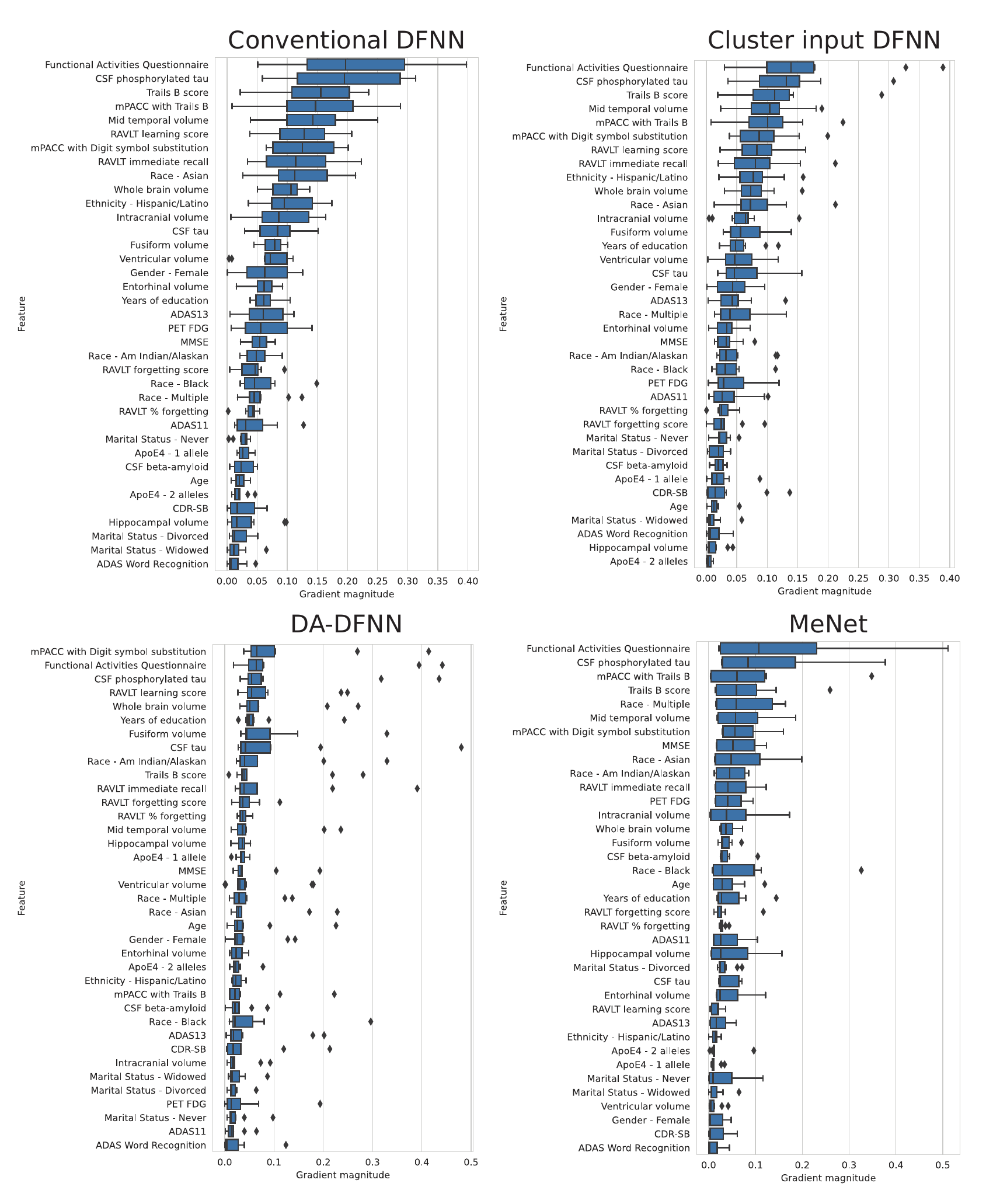}
    \caption{Feature importance in classification of stable vs. progressive mild cognitive impairment with various DFNN models. Features are ranked by descending median feature importance (gradient magnitude) across 10 cross-validation folds.}
    \label{fig:mci_features_all_models}
\end{figure}

\begin{figure} \ContinuedFloat
    \centering
    \includegraphics[width=\textwidth]{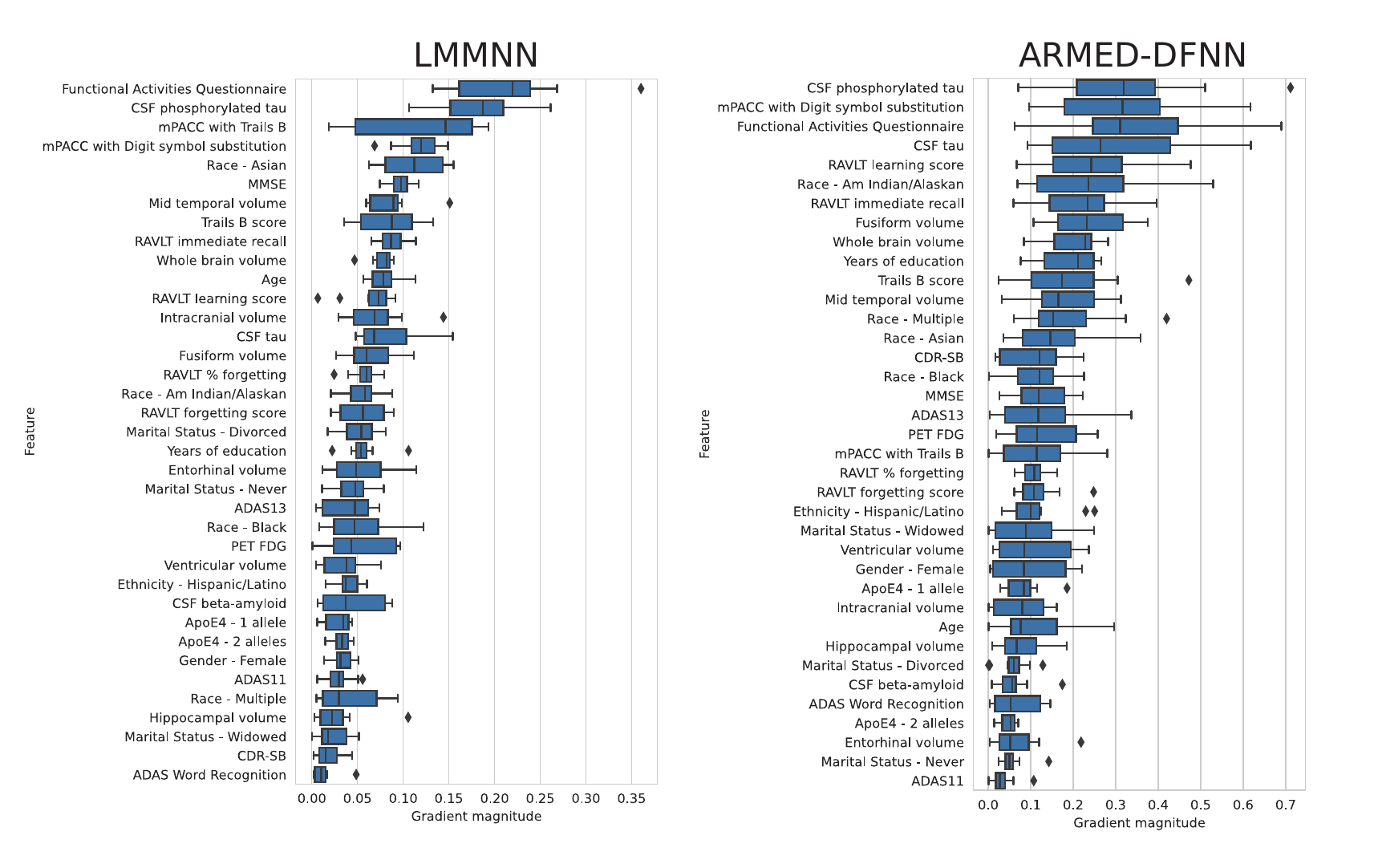}
    \caption{Feature importance in the classification of stable vs. progressive mild cognitive impairment with various DFNN models (continued).}
\end{figure}

\begin{figure}[h]
    \centering
    \includegraphics[width=\textwidth]{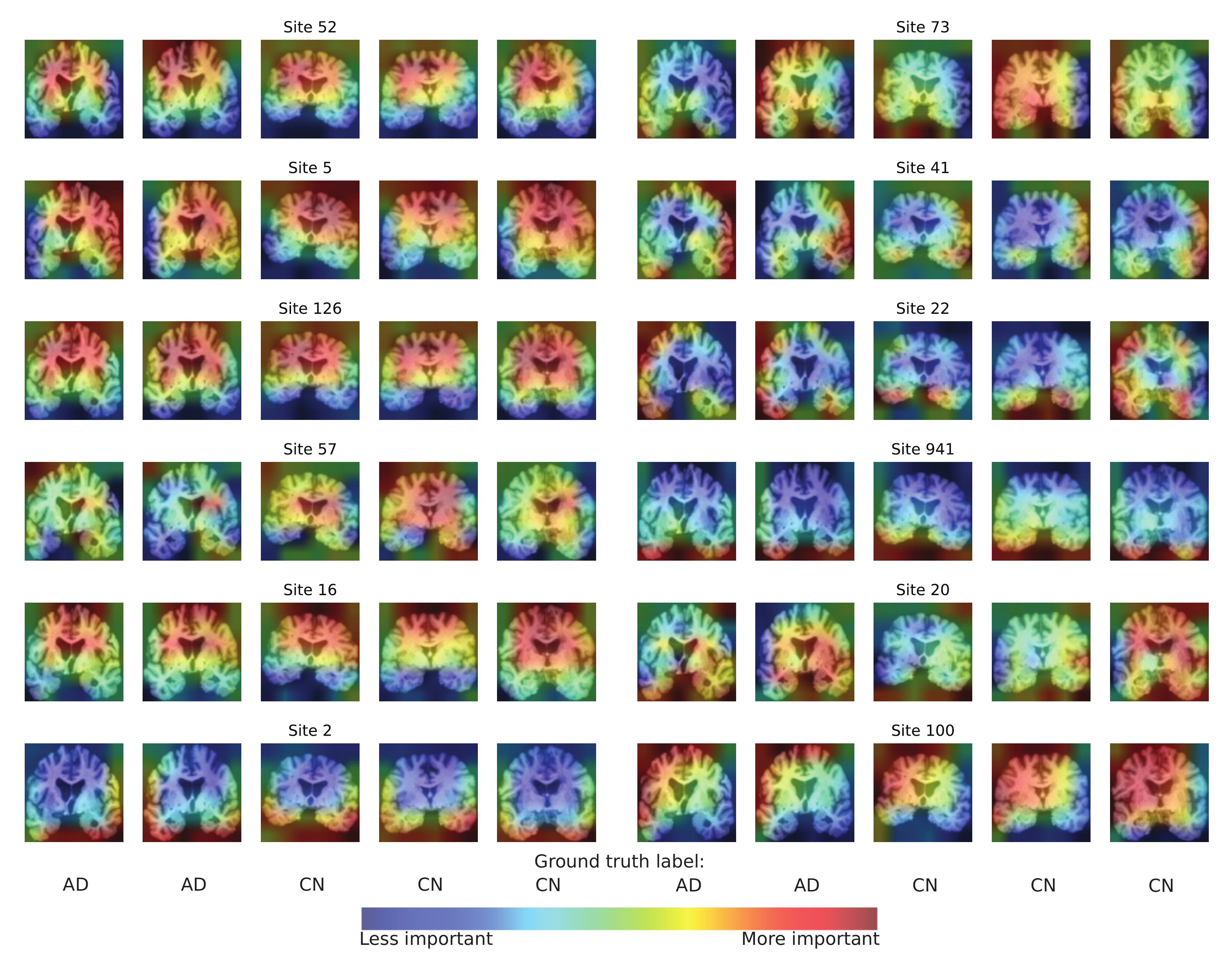}
    \caption{Site-specific random effects involved in Alzheimer's Disease diagnosis, as learned by the ARMED-CNN. For the same representative 5 images, we computed two Grad-CAM visualizations for each site $j$: $G_F$ using the fixed effects subnetwork only and $G_M(j)$ using the full model applying the specific random effects for site $j$. Shown here are the difference images ($G_F - G_M(j)$) which reveal the contribution of the random effects subnetwork in predicting diagnosis. Colormaps are scaled per image. Ground truth labels, Alzheimer's Disease (AD) or cognitively normal (CN), are indicated at the bottom.}
    \label{fig:gradcam_re}
\end{figure}

\newpage
\section{Supplemental methods}

\subsection{Specific mixed effects model architectures}
\subsubsection{Dense feedforward neural network (DFNN)} \label{supmethods:dfnn}
Our base DFNN architecture for both the spiral classification and the MCI conversion prediction problems contained 3 hidden dense layers with 4 neurons each (Fig. \ref{fig:dfnn_architecture}, blue area). Each was followed by a ReLU activation function. For binary classification, the output layer was a single-neuron dense layer with a sigmoid activation. To construct the ARMED-DFNN, we began by introducing an adversarial classifier containing 3 hidden dense layers with 8, 8, and 4 neurons, respectively (conceptualized in Fig. \ref{fig:dfnn_architecture}, gray area). ReLU activations were used for the hidden layers, and the output layer was a softmax layer for multi-class classification. The random effects subnetwork (Fig. \ref{fig:dfnn_architecture}, orange area) of the ARMED-DFNN varied by problem. For the \textit{spiral classification} problem where the random effect was nonlinear by design, we included an 4-dimensional cluster-specific nonlinear slope (main text Eq. 4), which was multiplied with the output of the last hidden dense layer in the main (fixed effects) subnetwork, and a cluster-specific intercept (main text Eq. 5):
\begin{equation} \label{eq:re_nonlinear_intercept}
    h_R(\boldsymbol{x}_i; U(\boldsymbol{z}_i)) = h_{R, nlin}(\boldsymbol{x}_i; u_{nlin}(\boldsymbol{z}_i)) + h_{R, int}(u_{int}(\boldsymbol{z}_i))
\end{equation}
For the \textit{MCI conversion prediction} problem, we included a $p$-dimensional cluster-specific linear slope (main text Eq. 6), which was multiplied by the input $X$ with $p$ features, and a cluster-specific intercept (main text Eq. 5):
\begin{equation} \label{eq:re_linear_intercept}
    h_R(\boldsymbol{x}_i; U(\boldsymbol{z}_i)) = h_{R, lin}(\boldsymbol{x}_i; u_{lin}(\boldsymbol{z}_i)) + h_{R, int}(u_{int}(\boldsymbol{z}_i))
\end{equation}
To combine the fixed ($h_F(\boldsymbol{x}_i; \beta)$) and random effects ($h_R(\boldsymbol{x}_i; U(\boldsymbol{z}_i))$) subnetwork outputs and obtained the mixed effects prediction $\hat{y}_{M, i}$, we used the following additive mixing function:
\begin{equation} \label{eq:me_mixing1}
    \hat{y}_{M, i} = m \left(h_F(\boldsymbol{x}_i; \beta), h_R(\boldsymbol{x}_i; U(\boldsymbol{z}_i)) \right) = sigmoid \left(h_F(\boldsymbol{x}_i; \beta)^\top \beta_L + h_R(\boldsymbol{x}_i; U(\boldsymbol{z}_i)) \right)
\end{equation}
where $\beta_L$ contains the weights of the output layer in the fixed effects subnetwork. The final training objective for the ARMED-DFNN is
\begin{equation} \label{eq:me_dfnn_loss}
    \mathcal{L}_{BCE}(\boldsymbol{y}, \hat{\boldsymbol{y}}_M) + \lambda_{F} \mathcal{L}_{BCE}(\boldsymbol{y}, \hat{\boldsymbol{y}}_F) + \lambda_{K} D_{\text{KL}}(q(U) || p(U)) - \lambda_g \mathcal{L}_{CCE}(Z, \hat{Z}) 
\end{equation}
where $\mathcal{L}_{BCE}$ is the binary crossentropy loss:
\begin{equation*}
    \mathcal{L}_{BCE}(\boldsymbol{y}, \hat{\boldsymbol{y}}) = - \frac{1}{n} \sum_{i=1}^n y_i \log (\hat{y}_i) + (1 - y_i) \log (1 - \hat{y}_i)
\end{equation*}

The Z-predictor model, which infers the cluster membership matrix $Z$ of unseen sites, used the same architecture as the adversarial classifier. All DFNN models were trained with the Adam optimizer with a learning rate of 0.001 for 50 epochs, with early stopping based on validation loss.

\subsubsection{Convolutional neural network (CNN)} \label{supmethods:cnn}
The base CNN used for AD vs. CN classification contained 7 blocks, each comprising a two-dimensional convolutional layer with $3 \times 3$ kernels, batch normalization, and a PReLU activation function (Fig. \ref{fig:cnn_architecture}, blue area). A $2 \times 2$ max-pooling layer was used between each block. The output of the last convolutional block was flattened and fed into a 512-neuron dense layer. Here, $h_F(X; \beta)$ is defined as the output of this hidden dense layer. The output layer was a single-neuron dense layer with a sigmoid activation. To form the fixed effects subnetwork of the ARMED-CNN, we first added an adversarial classifier (Fig. \ref{fig:cnn_architecture}, gray area). We used an architecture similar to the base CNN, but the output layer was replaced by a softmax layer for multi-class classification. Intermediate outputs from each layer in the original CNN were sent into the adversarial classifier at the layer with the corresponding shape. Next, the random effects subnetwork (Fig. \ref{fig:cnn_architecture}, orange area) consisted of a cluster-specific bias and 512-dimensional cluster-specific scalars (nonlinear slopes, akin to Eq. \ref{eq:re_nonlinear_intercept}), which was multiplied with $h_F(X; \beta)$. We used the mixing function in Eq. \ref{eq:me_mixing1} to combine the fixed and random effects subnetwork outputs. The loss function was the same as in Eq. \ref{eq:me_dfnn_loss}. The Z-predictor model used the same architecture as the adversarial classifier and all models were trained with the Nadam optimizer with a learning rate of 0.0001 for 20 epochs. 

\subsubsection{Autoencoder-classifier (AEC)} \label{supmethods:aec}
We began with the autoencoder architecture used in \cite{Zaritsky.2021} (Fig. \ref{fig:aec_architecture}, blue area), which contains an encoder to compress an image into a 56-dimensional latent representation $e_F(X; \beta)$. A decoder then reconstructs the image $\hat{X}_F$. The encoder contained 6 blocks, each with a convolutional layer with $4 \times 4$ kernels and $2 \times 2$ striding, batch normalization, and PReLU activation. The output of the last convolutional block was flattened and fed into a 56-neuron dense layer with produced the compressed latent representation. The decoder architecture was symmetric and replaced convolutional layers with transposed convolutional layers. To simultaneously perform classification, we introduced an auxiliary classifier subnetwork which predicts the cell phenotype $\hat{\boldsymbol{y}}_F$, i.e. high vs. low metastatic efficiency, from the latent representation. The auxiliary classifier took the encoder's latent representation as input and contained a 32-neuron hidden layer and a sigmoid output layer.

To create the fixed effects subnetwork, we added an adversarial classifier to predict each image's cluster $\hat{Z}_A$ from the layer activations of the encoder (Fig. \ref{fig:aec_architecture}, gray area). Through the generalization loss, the autoencoder is penalized for learning features that allow accurate cluster prediction. The encoder and decoder weights are tied so that this penalty affects both modules of the autoencoder. The adversary used the same architecture as the encoder, with the addition of a final softmax output layer. 

Because image-level batch effects are believed to pervade all levels of the feature hierarchy, including lower level pixel-level features as well as higher-level morphology features, we construct a second, mirrored autoencoder as the random effects subnetwork (Fig. \ref{fig:aec_architecture}, orange area). In this second autoencoder, called the RE-AEC, random effect features are learned at every layer using random effects convolution (RECON) blocks (Fig. \ref{fig:recon_block}), which were inspired by style transfer networks \cite{Dumoulin.10242016, Ulyanov.2017}. RECON blocks replaced the batch normalization layers of the original autoencoder. After a convolutional layer, these blocks apply instance normalization. In contrast to batch normalization which normalizes each mini-batch to zero mean and unit variance, instance normalization normalizes each sample independently \cite{Ulyanov.2017}. Afterward, a cluster-specific random scale and bias is applied to each feature map, effectively rescaling and shifting the convolution outputs in a cluster-dependent manner. 

The RE-AEC produces a cluster effect-laden latent representation $e_R(X; U(Z))$ and reconstruction $\hat{X}_R$. To further enforce the learning of cluster-specific features in the RE-AEC, we added two more classifiers to predict cluster membership from the latent representation, $\hat{Z}_L$, and from the image reconstruction, $\hat{Z}_I$. By minimizing the prediction error of these two classifiers, the RE-AEC is encouraged to produce latent representations and reconstructions characterizing the cluster effects. 

To create the ARMED-AEC, we combined the fixed effects subnetwork (the domain adversarial AEC, denoted as DA-AEC) with the RE-AEC. Specifically, we concatenated the latent representations $e_F(X; \beta)$ and $e_R(X; U(Z))$ into a single vector and used this as input to a separate DFNN classifier trained to produce a mixed effects-based phenotype prediction $\hat{\boldsymbol{y}}_M$. In other words, the mixing function was
\begin{equation*}
    \hat{y}_{M, i} = m \left(h_F(\boldsymbol{x}_i; \beta), h_R(\boldsymbol{x}_i; U(\boldsymbol{z}_i)) \right) = m([h_F(\boldsymbol{x}_i; \beta), h_R(\boldsymbol{x}_i; U(\boldsymbol{z}_i)]) 
\end{equation*}
where $m(...)$ represents this new classifier using the concatenated latent representations as input. The combined training objective of the ARMED-AEC was
\begin{equation} \label{eq:aec_loss}
    \begin{split}
        &\lambda_{recon,F} \mathcal{L}_{MSE}(X, \hat{X}_F) + \lambda_{recon,R} \mathcal{L}_{MSE}(X, \hat{X}_R) \\
        & + \lambda_{class,F} \mathcal{L}_{BCE}(\boldsymbol{y}, \hat{\boldsymbol{y}}_F) + \lambda_{class,M} \mathcal{L}_{BCE}(\boldsymbol{y}, \hat{\boldsymbol{y}}_M) \\
        & + \lambda_{K} D_{\text{KL}}(q(U) || p(U)) \\
        & + \lambda_{L} \mathcal{L}_{CCE}(Z, \hat{Z}_L)  + \lambda_{I} \mathcal{L}_{CCE}(Z, \hat{Z}_I) \\
        & - \lambda_g \mathcal{L}_{CCE}(Z, \hat{Z}_A) 
    \end{split}
\end{equation}
The first line in Eq. \ref{eq:aec_loss} contains the image reconstruction loss for the fixed and random effects subnetworks. Here, we use the mean squared error (MSE) between the input $X$ and the reconstruction $\hat{X}$. The second line contains the phenotype classification loss (binary crossentropy) between the true and predicted phenotype labels $\boldsymbol{y}$ and $\hat{\boldsymbol{y}}$. The third line is the KL divergence for Bayesian layers. The fourth line contains the categorical crossentropy between the true cluster label $Z$ and the cluster labels predicted by the latent-cluster classifier $\hat{Z}_L$ and the image-cluster classifier $\hat{Z}_{I}$. Finally, the last line is the cluster generalization loss which encourages the fixed effects subnetwork to learn features that prevent the adversary from predicting cluster labels $\hat{Z}_A$. 

We used an additional CNN classifier analogous to the adversarial classifier as the Z-predictor. All AEC models were trained with the Nadam optimizer with a learning rate of 0.0001 for 20 epochs, with early stopping based on reconstruction mean squared error on validation data.  

\subsection{Spiral classification simulations} \label{supmethods:spirals_dataset}
The spiral simulations generated two spirals, labeled $y = 0$ and $y = 1$ and separated in phase by $\pi$ radians (see Fig. \ref{fig:spiral_data} and Eq. \ref{eq:spirals}). We randomly generated $n = 10,000$ points along the spirals at arc lengths between 0 and $2\pi$. Consequently, the measured features for the $i^{th}$ point are its two-dimensional coordinates $x_{i, 1}$ and $x_{i, 2}$ and the label is $y_i$. We then added Gaussian noise $\eta \sim N(0, 0.1)$ to $X$ to increase the classification challenge. 

\begin{equation} \label{eq:spirals}
    \begin{aligned} 
        x_{i, 1} & = - \frac{r_j t_i}{2\pi} \cos(t_i - \phi) + \eta\\
        x_{i, 2} & = \frac{r_j t_i}{2\pi} \sin(t_i - \phi) + \eta\\
        t & \in \mathbb{R}^{n \times 1} \sim \text{Uniform}(0, 2\pi) \\
        \phi & = \begin{cases}
                0, & y = 0 \\
                \pi, & y = 1 
        \end{cases}
    \end{aligned}
\end{equation}
\ignorespacesafterend

To simulate random effects, we divided the points evenly into $c = 10$ clusters and varied the spiral radius $r_j$ for each cluster $j$. In simulation 1, we sampled the cluster-specific radii from a normal distribution centered at 1, $r_j \sim N(1, 0.3)$ (Fig. \ref{fig:spiral_data}a). In simulation 2, we increased the severity of the random perturbations by sampling the cluster-specific radii from a normal distribution centered at 0, $r_j \sim N(0, 1.0)$, which causes the spirals to be inverted in half of the clusters (Fig. \ref{fig:spiral_data}b). In simulation 3, we added two new features $x_3$ and $x_4$ which were confounded, i.e. they were associated with spiral label, $y$, but uncorrelated with the underlying spiral function. To simulate confounded variables, we first sampled a random variable $\rho_j \sim N(0, 0.1)$ for each cluster. We let $\rho_j$ determine the class balance for each cluster, e.g. in cluster 3 with $\rho_3 = 0.2$, 20\% of samples belong to class $y = 0$ and 80\% belong to class $y = 1$. We then generated two additional data features as a function of $\rho_j$ (Eq. \ref{eq:spiralProbes}), thereby creating a confounded relationship between $x$ and $y$ that is unrelated to the true underlying spiral function (Fig. \ref{fig:spiral_data}c).
\begin{equation} \label{eq:spiralProbes}
    \begin{aligned}
        x_{i, 3} & \sim N(\rho_j, 0.2) \\
        x_{i, 4} & \sim N(\rho_j, 0.2)
    \end{aligned}
\end{equation}
\ignorespacesafterend

\subsection{Alzheimer's Disease and mild cognitive impairment clincal and neuroimaging datasets}
Data used in the preparation of this article were obtained from the Alzheimer’s Disease Neuroimaging
Initiative (ADNI) database (\url{https://adni.loni.usc.edu}). The ADNI was launched in 2003 as a public-private
partnership, led by Principal Investigator Michael W. Weiner, MD. The primary goal of ADNI has been to
test whether serial magnetic resonance imaging (MRI), positron emission tomography (PET), other
biological markers, and clinical and neuropsychological assessment can be combined to measure the
progression of mild cognitive impairment (MCI) and early Alzheimer’s disease (AD).

\subsubsection{MCI conversion prediction dataset} \label{supmethods:mci_dataset}
The DFNN models for the classification of progressive vs. stable MCI (pMCI vs. sMCI) were trained on the curated ``ADNIMERGE" dataset provided by ADNI. This dataset contains a selection of key variables from the ADNI study. We selected the subjects with the MCI diagnosis at the baseline visit and who had a 24-month follow-up visit. We then selected the 20 sites with the most subjects and held out the remaining 34 sites as the ``unseen site" testing data. Subjects were labeled pMCI if their 24-month diagnosis changed to Alzheimer's Disease and sMCI otherwise. The 20 ``seen sites" contained 392 subjects including 106 (27\%) pMCI subjects and 286 (73\%) sMCI subjects. We performed $10 \times 10$ fold nested cross-validation to optimize model hyperparameters and we report test performance on the outer folds in the results. The unseen sites contained 313 subjects including 123 (38\%) pMCI subjects and 190 (62\%) sMCI subjects. We selected 37 features with measurements available in at least 70\% of the subjects. These included:
\begin{itemize}
    \item Demographics: gender, ethnicity, race, marital status, years of education, and age
    \item Cognitive scores: Clinical Dementia Rating - Sum of Boxes (CDR-SB), Alzheimer's Disease Assessment Scale - Cognitive Subscale, 11-item  and 13-item versions (ADAS11, ADAS13), ADAS word recognition score, Mini Mental State Exam (MMSE), Rey Auditory Verbal Learning Test (RAVLT) subscores, Trail Making Test - Part B score, Functional Activities Questionnaire, modified Preclinical Alzheimer's Cognitive Composite (mPACC) with digit symbol substitution score and with Trails B score
    \item Region volumes from structural MRI: ventricles, hippocampus, whole brain, entorhinal cortex, fusiform gyrus, medial temporal lobe, and intracranial volume
    \item Fluorodeoxyglucose PET: average of the mean regional standardized uptake values (SUVr) from each of the following regions: angular gyrus, temporal lobe, and posterior cingulate gyrus 
    \item Cerebrospinal fluid concentrations: tau, phosphorylated tau, beta-amyloid
    \item Number of APOE4 alleles
\end{itemize}

\paragraph{Addition of confounded probes to the ADNI data} \label{supmethods:mci_probes}
We simulated 5 confounded probe features to test each model's ability to discern confounding effects which are spuriously correlated with the prediction target. First, we computed the percentage of pMCI subjects $\rho_j$ for each site $j$. We then generated the following 5 features as linear and nonlinear functions of $\rho_j$:
\begin{align*}
    x_1 &= \rho_j + \eta \\
    x_2 &= \rho_j ^ 2 + \eta \\
    x_3 &= \frac{1}{\rho_j} + \eta \\
    x_4 &= \cos{\rho_j} + \eta \\
    x_5 &= \sin{\rho_j} + \eta \\
\end{align*}
where $\eta$ is Gaussian noise with $\eta \sim N(0, 0.05)$. Consequently, these probe features were associated with the prediction target, MCI conversion, without having any biological relevance. In the experiments, we evaluate the sensitivity of each model to these probes.

\subsubsection{AD diagnosis dataset} \label{supmethods:ad_dataset}
The data for training CNN classifiers of AD vs. cognitively normal (CN) individuals was selected from the ADNI2 and ADNI3 cohorts. To emphasize the underlying confounding site effect in this dataset, we selected 12 study sites that differed in scanner manufacturer, scanner model, and the proportion of AD samples (Table \ref{table:adni_scanners}). This included 5 sites using General Electric scanners with a high proportion of AD samples ($\geq$ 50\%) and 7 sites using Philips or Siemens scanners with a low proportion of AD samples ($\leq$ 11.3\%). Consequently, there is a strong confounding effect due to the artifactual association between image acquisition characteristics and the likelihood of an AD diagnosis. 

We obtained 2D coronal slices as follows. First, T1-weighted MRI were skullstripped using CONSNet \cite{Lucena.2019}. The results were manually inspected and images with poor skullstripping were discarded. Next, a nonlinear transformation between the skullstripped images and the Montreal Neurological Institute (MNI) ICBM152 brain template was computed using ANTs \cite{Avants.2011, Tustison.2014, Fonov.2009}. We selected a point in the right hippocampus at MNI coordinate $(x=32, y=-6, z=-26)$. We transformed this coordinate back into subject space using the inverse transformation from ANTs and extracted the coronal slice orthogonal to the anterior-posterior axis in subject space and passing through this coordinate. Finally, we normalized image intensity by scaling the 2nd and 98th percentiles of the intensity histogram to 0 and 255, respectively. 

This resulted in a final dataset containing the 12 ``seen sites" included 482 images from 184 subjects. Given this sample size, we chose to perform Monte Carlo cross-validation with 10 random splits instead of K-fold cross-validation. For each split, the subjects were divided into 70\% training, 10\% validation, and 20\% test partitions, with all images from each subject assigned to the same partition. Model hyperparameters were optimized based on validation performance and final test performance on the 12 seen sites is reported in the results. The ``unseen sites" dataset contained 1,703 images from 673 subjects, from the remaining 51 sites, which was used to evaluate the performance of the models on sites not used during training. 

\subsection{Melanoma live-cell image dataset} \label{supmethods:melanoma_dataset}
Complete details of this dataset can be found in Zaritsky et al. 2021 \cite{Zaritsky.2021}. Here, we briefly describe the data included in the current analysis. We selected images from 7 patient-derived xenografts (PDXs) transplanted from melanoma patients to immunocompromised mice. These included 4 PDXs categorized as low metastatic efficiency, meaning the patient had metastases only to the lung and was successfully treated, and 3 PDXs categorized as high metastatic efficiency, meaning the patient had widespread metastases and died. Cells from each PDX were placed on collagen plates and recorded with phase contrast microscopy, at a rate of 1 image per minute for 2 hours, over 24 days (i.e. batches). Not every PDX was imaged in each batch, so we selected 13 batches where cells from at least 2 PDXs were imaged to constitute the ``seen batches" dataset. This contained 120,863 images from 3,260 cells. We divided this dataset into 70\% training, 10\% validation, and 20\% test partitions, with all images from the same cell assigned to the same partition. Performance on the test partition is reported in the results. The ``unseen batches" dataset contained 10,699 images from 2,351 cells, imaged over the remaining 11 batches. 

\bibliographystyle{IEEEtran}
\bibliography{references}